\documentclass[twoside,11pt]{article}
\usepackage{times} 
\usepackage{theapa}
\usepackage{amsmath}
\usepackage{amsthm}
\usepackage{amssymb}
\usepackage{epsf}
\usepackage{glossary}
\usepackage{jair}                                                               
\usepackage[dvips]{graphicx}
\usepackage{latexsym}
\usepackage{xspace}

\usepackage{algorithm}
\usepackage{algorithmic}

\renewcommand{\(}{\left(}
\renewcommand{\)}{\right)}
\renewcommand{\[}{\left[}
\renewcommand{\]}{\right]}

\newcommand{\eg}{{\em e.g.\ }}

\newcommand{\U}{{\cal U}}

\newcommand{\Y}{{\cal Y}}

\newcommand{\calS}{{\cal S}}

\DeclareMathOperator{\sign}{sign}

\DeclareMathOperator{\TD}{TD}

\DeclareMathOperator{\GPOMDP}{GPOMDP}
\DeclareMathOperator{\CONJGRAD}{CONJPOMDP}
\DeclareMathOperator{\OLPOMDP}{OLPOMDP}

\DeclareMathOperator{\MDP}{MDP}
\DeclareMathOperator{\POMDP}{POMDP}

\DeclareMathOperator{\GSEARCH}{GSEARCH}

\DeclareMathOperator{\GRAD}{GRAD}

\newcommand{\pomdpg}{{\mbox{\small$\GPOMDP$}}}
\newcommand{\gpomdp}{{\mbox{\small$\GPOMDP$}}}

\newcommand{\conjgrad}{{\mbox{\small$\CONJGRAD$}}}
\newcommand{\olpomdp}{{\mbox{\small$\OLPOMDP$}}}

\newcommand{\mdp}{{\mbox{\small$\MDP$}}}

\newcommand{\pomdp}{{\mbox{\small$\POMDP$}}}
\newcommand{\pomdps}{{\mbox{\small$\POMDP$}}s}
\newcommand{\gsearch}{{\mbox{\small$\GSEARCH$}}}

\newcommand{\grad}{{\mbox{\small$\GRAD$}}}

\newcommand{\nb}{{\nabla_{\!\!\beta}}}
\newcommand{\Expect}{\mathbold{E}}

\newcommand{\E}{\Expect}
\newcommand{\mathbold}[1]{\mbox{\boldmath $\bf#1$}}

\newcommand{\R}{{\mathbb R}}

\usepackage{color}

\definecolor{darkred}{rgb}{0.7,0.2,0.2}

\definecolor{bgblue}{rgb}{0.04,0.39,0.53}

\newtheorem{assumption}{Assumption}

\begin{document}
\title{Experiments with Infinite-Horizon, Policy-Gradient Estimation}
\author{\name Jonathan Baxter \\
\email jbaxter@whizbang.com \\
\addr{WhizBang! Labs.}\\
\addr{4616 Henry Street Pittsburgh, PA  15213} \\
\name Peter L. Bartlett
\email bartlett@barnhilltechnologies.com\\
\addr{BIOwulf Technologies.} \\
\addr{2030 Addison Street, Suite 102, Berkeley,CA 94704}\\
\name Lex Weaver \email Lex.Weaver@anu.edu.au \\
\addr{Department of Computer Science} \\
\addr{Australian National University , Canberra 0200, Australia}}
\firstpageno{351}
\jairheading{15}{2001}{}{9/00}{10/01}
\maketitle
\ShortHeadings{Policy-Gradient Estimation}{Baxter et al}
\begin{abstract}

%PB Tried to make the abstract less a review of the first paper.

In this paper, we present algorithms that perform gradient ascent of
the average reward in a partially observable Markov decision process
(\pomdp).  These algorithms are based on \pomdpg, an algorithm introduced
in a companion paper \cite{jair_01a}, which computes biased estimates of
the performance gradient in \pomdps.  The algorithm's chief advantages
are that it uses only one free parameter $\beta\in [0,1)$, which has a
natural interpretation in terms of bias-variance trade-off, it requires
no knowledge of the underlying state, and it can be applied to infinite
state, control and observation spaces.  We show how the gradient estimates
produced by \gpomdp\ can be used to perform gradient ascent, both with a
traditional stochastic-gradient algorithm, and with an algorithm based on
conjugate-gradients that utilizes gradient information to bracket maxima
in line searches. Experimental results are presented illustrating both
the theoretical results of \citeA{jair_01a} on a toy problem, and practical
aspects of the algorithms on a number of more realistic problems.

\end{abstract}

\section{Introduction}
\label{section:intro}

Function approximation is necessary to avoid the curse of
dimensionality associated with large-scale dynamic programming
and reinforcement learning problems. The dominant paradigm is to
use the function to approximate the state (or state and action)
values. Most algorithms then seek to minimize some form of error
between the approximate value function and the true value function,
usually by simulation \cite{sutton98,bertsekas96}. While there have
been a multitude of empirical successes for this approach \cite<for
example,>{samuel59,tesauro92,tesauro94,ml_00a,zhang95,singh97}, there
are only weak theoretical guarantees on the performance of the policy
generated by the approximate value function. In particular, there is no
guarantee that the policy will improve as the approximate value function
is trained; in fact performance can degrade even when the function class
contains an approximate value function whose corresponding greedy policy
is optimal \cite<see>[Appendix A, for a simple two-state example]{jair_01a}.

An alternative technique that has received increased attention
recently is the ``policy-gradient'' approach in which the parameters
of a stochastic policy are adjusted in the direction of the gradient
of some performance criterion (typically either expected
discounted reward or average reward). The key problem is how to
compute the performance gradient under conditions of partial
observability when an explicit model of the system is not
available. 

This question has been addressed in a large body of previous work
\cite{BarSutAnd83,williams92,glynn86,cao97,cao98,fu94,singh94a,singh95,marbach98,marbachthesis98,baird98,rubinstein98,kimura95,kimura97}.
See the introduction of \cite{jair_01a} for a discussion of the history of
policy-gradient approaches. Most existing algorithms rely on the existence
of an identifiable recurrent state in order to make their updates to the
gradient estimate, and the variance of the algorithms is governed by the
recurrence time to that state. In cases where the recurrence time is too
large (for instance because the state space is large), or in situations of
partial observability where such a state cannot be reliably identified,
we need to seek alternatives that do not require access to such a state.

Motivated by these considerations, \citeA{jair_01a,icml_00} introduced and
analysed \gpomdp---an algorithm for generating a {\em biased} estimate
of the gradient of the average reward in general Partially Observable
Markov Decision Processes (\pomdps) controlled by parameterized stochastic
policies.
%PB We don't know this.
% \footnote{It seems that without access to an
% underlying state, any general algorithm for estimating the performance
% gradient must necessarily be biased.}
The chief advantages of \gpomdp\ are that it requires only a single
sample path of the underlying Markov chain, it uses only one free
parameter $\beta\in [0,1)$, which has a natural interpretation in
terms of bias-variance trade-off, and it requires no knowledge of the
underlying state.
%PB This is an incomplete description - I think it's best to rely on
%PB the citations and pointer to history in previous para.
% \gpomdp\ is essentially an extension of Williams' \reinforce\
% algorithm \cite{williams92} and similar more recent algorithms
% \cite{kimura97,cao98,marbach98}.

More specifically, suppose $\theta\in\R^K$ are the parameters
controlling the \pomdp. For example, $\theta$ could be the parameters
of an approximate neural-network value-function that generates a
stochastic policy by some form of randomized look-ahead, or $\theta$
could be the parameters of an approximate $Q$ function used to
stochastically select controls\footnote{Stochastic policies are not
strictly necessary in our framework, but the policy must be
``differentiable'' in the sense that $\nabla \eta(\theta)$
exists.}. Let $\eta(\theta)$ denote the average reward of the \pomdp\
with parameter setting $\theta$.  \pomdpg\ computes an approximation
$\nb \eta(\theta)$ to $\nabla \eta(\theta)$ based on a single
continuous sample path of the underlying Markov chain. The accuracy of
the approximation is controlled by the parameter $\beta\in [0,1)$, and
one can show that 
$$
\nabla\eta(\theta) = \lim_{\beta\rightarrow 1} \nb\eta(\theta).
$$ 
The trade-off preventing choosing $\beta$ arbitrarily close to 1 is
that the {\em variance} of \pomdpg's estimates of $\nb\eta(\theta)$ scale as
$1/(1-\beta)^2$. However, on the bright side, it can also be shown
that the {\em bias} of $\nb(\theta)$ (measured by $\|\nb\eta(\theta) -
\nabla\eta(\theta)\|$) is proportional to $\tau(1-\beta)$ where $\tau$ is a
suitable {\em mixing time} of the Markov chain underlying the \pomdp\
\cite{colt_00}.  Thus for ``rapidly mixing'' \pomdp's (for which
$\tau$ is small), estimates of the performance gradient with
acceptable {\em bias} and {\em variance} can be obtained.

Provided $\nb \eta(\theta)$ is a sufficiently accurate approximation
to $\nabla \eta(\theta)$---in fact, $\nb \eta(\theta)$ need only be
within $90^\circ$ of $\nabla \eta(\theta)$---small adjustments to the
parameters $\theta$ in the direction $\nb \eta(\theta)$ will guarantee
improvement in the average reward $\eta(\theta)$. In this case,
gradient-based optimization algorithms using $\nb \eta(\theta)$ as
their gradient estimate will be guaranteed to improve the average
reward $\eta(\theta)$ on each step. Except in the case of
table-lookup, most value-function based approaches to reinforcement
learning cannot make this guarantee.

In this paper we present a conjugate-gradient ascent algorithm that
uses the estimates of $\nb \eta(\theta)$ provided by \pomdpg. Critical
to the successful operation of the algorithm is a novel line search
subroutine that brackets maxima by relying solely upon gradient
estimates. This largely avoids problems associated with finding the
maximum using noisy value estimates. Since the parameters are only
updated after accumulating sufficiently accurate estimates of the
gradient direction, we refer to this approach as the ``off-line''
algorithm. This approach essentially allows us to take a stochastic
gradient optimization problem and treat it as a non-stochastic
optimization problem, thus enabling the use of a large body of
accumulated heuristics and algorithmic improvements associated with
such methods. We also present a more traditional, ``on-line''
stochastic gradient ascent algorithm based on
\pomdpg\ that updates the parameters at every time step. This
algorithm is essentially the algorithm proposed in
\cite{kimura97}. 

The off-line and on-line algorithms are applied to a variety of
problems, beginning with a simple 3-state Markov decision process
(MDP) controlled by a linear function for which the true gradient can
be exactly computed. We show rapid convergence of the gradient
estimates $\nb\eta(\theta)$ to the true gradient, in this case over a
large range of values of $\beta$. With this simple system we are able
to illustrate vividly the bias/variance tradeoff associated with the
selection of $\beta$. We then compare the performance of the off-line
and on-line approaches applied to finding a good policy for the
MDP. The off-line algorithm reliably finds a near-optimal policy in
less than 100 iterations of the Markov chain, an order of magnitude
faster than the on-line approach. This can be attributed to the more
aggressive exploitation of the gradient information by the off-line
method.

Next we demonstrate the effectiveness of the off-line algorithm in
training a neural network controller to control a ``puck'' in a
two-dimensional world. The task in this case is to reliably navigate
the puck from any starting configuration to an arbitrary target
location in the minimum time, while only applying discrete forces in
the $x$ and $y$ directions. Although the on-line algorithm was tried
for this problem, convergence was considerably slower and we were not
able to reliably find a good local optimum.

In the third experiment, we use the off-line algorithm to train a
controller for the call admission queueing problem treated in
\cite{marbachthesis98}.  In this case near-optimal
solutions are found within about 2000 iterations of the underlying
queue, 1-2 orders of magnitude faster than the experiments reported in
\cite{marbachthesis98} with on-line (stochastic-gradient) algorithms.

In the fourth and final experiment, the off-line algorithm was used to
reliably train a switched neural-network controller for a
two-dimensional variation on the classical ``mountain-car'' task
\cite[Example 8.2]{sutton98}.

%PB Combined the two advertisements in next section.

The rest of this paper is organized as follows. In
Section~\ref{sec:def} we introduce \pomdps\ controlled by stochastic
policies, and the assumptions needed for our algorithms to
apply. \pomdpg\ is described in Section ~\ref{sec:pomdpg}.  In
Section~\ref{sec:alg} we describe the off-line and on-line
gradient-ascent algorithms, including the gradient-based
line-search subroutine. Experimental results are presented in
Section~\ref{sec:exp}.

\section{$\mathbold{\pomdps}$\ Controlled by Stochastic Policies}
\label{sec:def}
A partially observable, Markov decision process (\pomdp) consists of a
state space $\calS$, observation space $\Y$ and a control space
$\U$. For each state $i\in \calS$ there is a deterministic reward
$r(i)$. Although the results in \citeA{jair_01a} only guarantee
convergence of \pomdpg\ in the case of finite $\calS$ (but continuous
$\U$ and $\Y$), the algorithm can be applied regardless of the nature
of $\calS$ so we do not restrict the cardinality of $\calS$, $\U$ or
$\Y$.

Consider first the case of discrete $\calS$, $\U$ and $\Y$.
Each control $u\in \U$ determines a stochastic matrix
$P(u) = [p_{ij}(u)]$ giving the transition probability from state
$i$ to state $j$ ($i,j\in \calS$).  For each state $i\in\calS$,
an observation $Y\in\Y$ is
generated independently according to a probability distribution $\nu(i)$
over observations in $\Y$.  We denote the probability that $Y=y$
by $\nu_y(i)$. A {\em randomized policy} is simply a function $\mu$ mapping
observations into probability distributions over the controls $\U$.
That is, for each observation $y\in \Y$, $\mu(y)$ is a distribution over the
controls in $\U$.  Denote the probability under $\mu$ of control $u$
given observation $y$ by $\mu_u(y)$.

For continuous $\calS, \Y$ and $\U$, $p_{ij}(u)$ becomes a {\em kernel}
$k_{ij}(u)$ giving the probability density of transitions from $i$
to $j$, $\nu(i)$ becomes a probability density function on $\Y$
with $\nu_y(i)$ the density at $y$, and $\mu(y)$ becomes a
probability density function on $\U$ with $\mu_u(y)$ the
density at $u$.

To each randomized policy $\mu$ there corresponds a Markov chain in
which state transitions are generated by first selecting an
observation $Y$ in state $i$ according to the distribution $\nu(i)$,
then selecting a control $U$ according to the distribution $\mu(Y)$,
and finally generating a transition to state $j$ according to the
probability $p_{ij}(U)$.

At present we are only dealing with a fixed \pomdp. To parameterize
the \pomdp\ we parameterize the policies,
so that $\mu$ now becomes a function $\mu(\theta, y)$ of a set of
parameters $\theta\in\R^K$, as well as of the observation $y$. The
Markov chain corresponding to $\theta$ has state transition matrix
$P(\theta) = [p_{ij}(\theta)]$ given by
\begin{equation}
\label{eq:barf}
p_{ij}(\theta) = \E_{Y\sim \nu(i)} \E_{U\sim \mu(\theta, Y)} p_{ij}(U).
\end{equation}
Note that the policies $\mu$ are {\em purely reactive} or {\em memoryless}
in that their choice of action is based only upon the current observation.
All the experiments described in the present paper use purely reactive
policies. \citeA{tr_belief_01} have extended \pomdpg\ and the techniques
of the present paper to controllers with internal state.

The following technical assumptions are required for the operation of
\pomdpg.
\begin{assumption}
\label{ass:deriv}
The derivatives,
$$
\frac{\partial \mu_u(\theta,y)}{\partial
\theta_k},
$$
exist, and the ratios 
$$
\dfrac{\left|\dfrac{\partial \mu_u(\theta,y)}{\partial
\theta_k}\right|}{\mu_u(\theta,y)}
$$
are uniformly bounded by $B < \infty$, for all $u\in \U$,
$y\in\Y$, $\theta \in \R^K$ and $k=1,\dots,K$. 
\end{assumption}
The second part of this assumption is needed because the ratio appears
in the \pomdpg\ algorithm. It allows zero-probability actions
$\mu_u(\theta, y) = 0$ only if $\nabla \mu_u(\theta,y)$ is also
zero, in which case we set $0/0 = 0$. See Section \ref{sec:exp} for
examples of policies satisfying this requirement. 
\begin{assumption}
\label{ass:rbound}
The magnitudes of the rewards, $|r(i)|$, are uniformly bounded by
$R < \infty$ for all states $i$.
\end{assumption}
For deterministic rewards, his condition only represents a
restriction in infinite state spaces. However, all the results in the
present paper apply to bounded stochastic rewards, in which case $r(i)$ is the
expectation of the reward in state $i$.
\begin{assumption}
\label{ass:station}
Each $P(\theta), \theta \in \R^K$, has a unique stationary
distribution $\pi(\theta)=\[\pi_1(\theta, \dots, \pi_n(\theta)\]$,
satisfying the {\em balance equations:}
$$
\pi(\theta) P(\theta) = \pi(\theta).
$$
\end{assumption}
Assumption \ref{ass:station} ensures that, for all parameters $\theta$,
the Markov chain forms a single recurrent class. Since any finite-state
Markov chain always ends up in a recurrent class, and it is the properties
of this class that determine the long-term average reward, this assumption
is mainly for convenience so that we do not have to include the recurrence
class as a quantifier in our theorems.  Observe that { \em episodic}
problems, such as the minimization of time to a goal state, may be modeled
in a way that satisfies Assumption~\ref{ass:station} by simply resetting
the agent upon reaching the goal state back to some initial starting
distribution over states. Examples are described in Section~\ref{sec:exp}.

The {\em average reward} $\eta(\theta)$ is simply the expected reward
under the stationary distribution $\pi(\theta)$:
\begin{equation}
\label{eq:avgreward}
\eta(\theta) = \sum_{i=1}^n \pi_i(\theta) r(i).
\end{equation}
Because of Assumption~\ref{ass:station}, 
$\eta(\theta)$ is also equal to the expected long-term average of the
reward received when starting from any state $i$:
$$
\eta(\theta) = \lim_{T\to\infty} \E\left(\left.\frac{1}{T}
  \sum_{t=0}^{T-1} r(X_t)\right| X_0=i\right).
$$
Here the expectation is over sequences of states $X_0,\ldots,X_{T-1}$
with state transitions generated by $P(\theta)$ (note that the
expectation is independent of the starting state $i$). 

\section{The $\mathbold{\pomdpg}$ Algorithm}
\label{sec:pomdpg}
\pomdpg\ (Algorithm \ref{algorithm:pgradmdp}) is an algorithm for
computing a {\em biased}
estimate $\Delta_T$ of the {\em gradient} of the average reward
$\nabla\eta(\theta)$.  $\Delta_T$ satisfies
$$
\lim_{T\rightarrow\infty} \Delta_T = \nb \eta(\theta),
$$
where $\nb\eta(\theta)$ ($\beta \in [0,1)$) is an approximation to
$\nabla\eta(\theta)$ satisfying
$$
\nabla\eta(\theta) = \lim_{\beta\rightarrow 1} \nb \eta(\theta),
$$
\cite[Theorems 2, 5]{jair_01a}. Note that \pomdpg\ relies only upon a single
sample path from the POMDP. Also, it does not
require knowledge of the transition probability matrix $P$, nor of the
observation process $\nu$; it only requires knowledge of the randomized
policy $\mu$, in particular the ability to compute the gradient of the
probability of the chosen control divided by the probability of the
chosen control. 

\begin{algorithm}
\caption{$\pomdpg(\beta, T, \theta) \rightarrow \R^K$ }%\cite[Algorithm 2]{jair_01a}.}
\label{algorithm:pgradmdp}
\begin{algorithmic}[1]
\STATE {\bf Given: }
\begin{itemize}
\item $\beta \in [0,1)$.
\item $T >0$.
\item Parameters $\theta\in \R^K$.
\item Randomized policy $\mu(\theta, \cdot)$
satisfying Assumption~\ref{ass:deriv}.
\item \pomdp\ with rewards satisfying Assumption~\ref{ass:rbound}, and
which when controlled by $\mu(\theta, \cdot)$ generates stochastic
matrices $P(\theta)$ satisfying Assumption~\ref{ass:station}.
\item Arbitrary (unknown) starting state $X_0$.
\end{itemize}
\STATE Set $z_0 = 0$ and $\Delta_0 = 0$ ($z_0, \Delta_0 \in\R^K$).
\FOR{$t=0$ to $T-1$}
\STATE Observe $Y_t$ (generated according to the observation
distribution $\nu(X_t)$)
\STATE Generate control $U_t$ according to $\mu(\theta, Y_t)$
\STATE Observe $r(X_{t+1})$ (where the next state $X_{t+1}$ is
generated according to $p_{X_tX_{t+1}}(U_t)$).
\STATE Set $z_{t+1} = \beta z_t +
  \dfrac{\nabla \mu_{U_t}(\theta, Y_t)}{\mu_{U_t}(\theta, Y_t)}$
\STATE Set $\Delta_{t+1} = \Delta_t + r(X_{t+1}) z_{t+1}$
\ENDFOR
\STATE $\Delta_T \leftarrow \Delta_T/T$
\STATE return $\Delta_T$
\end{algorithmic}
\end{algorithm}

We cannot set $\beta$ arbitrarily close to $1$ in \pomdpg, since the {\em
variance} of the estimate is proportional to $1/(1-\beta)^2$. However,
on the bright side, it can also be shown that the {\em bias} of
$\nb(\theta)$ (measured by $\|\nb\eta(\theta) - \nabla\eta(\theta)\|$)
is proportional to $\tau(1-\beta)$ where $\tau$ is a suitable {\em mixing
time} of the Markov chain underlying the \pomdp\ \cite{colt_00}.
Under Assumption \ref{ass:station}, regardless of the initial starting
state, the distribution over states converges to the stationary
distribution $\pi(\theta)$ when the agent is following policy $\mu(\theta,
\cdot)$. Standard Markov chain theory shows that the rate of convergence
to $\pi(\theta)$ is exponential, and loosely speaking, the mixing time
$\tau$ is the time constant in the exponential decay.

Thus $\beta$ has a natural interpretation in terms of a bias/variance
trade-off: small values of $\beta$ give lower variance in the estimates
$\Delta_T$, but higher bias in that the expectation of $\Delta_T$ may be
far from $\nabla \eta(\theta)$, whereas values of $\beta$ close to $1$
yield small bias but correspondingly larger variance. Fortunately,
for problems which mix rapidly (small $\tau$), $\beta$ can be
small and still yield reasonable bias. This bias/variance trade-off
is vividly illustrated in the experiments of Section~\ref{sec:exp};
see \cite{colt_00} for a more detailed theoretical discussion of the
bias/variance question.

\section{Stochastic Gradient Ascent Algorithms}
\label{sec:alg}

This section introduces two approaches to exploiting the gradient
estimates produced by \gpomdp: 
\begin{enumerate}
\item 
an off-line approach based on
traditional conjugate-gradient optimization techniques but employing a
novel line-search mechanism to cope with the noise in \gpomdp's
estimates, and
\item 
an on-line stochastic optimization approach that uses the core update in
\gpomdp\ ($r(X_t) z_t$) to update the parameters $\theta$ on {\em every}
iteration of the \pomdp.
\end{enumerate}

\subsection{Off-line optimization of the average reward}
\gpomdp\ generates biased 
and noisy estimates $\Delta_T$ of the gradient of the average reward
$\nabla\eta(\theta)$ for \pomdps\ controlled by parameterized
stochastic policies. A straightforward algorithm for finding local
maxima of $\eta(\theta)$ would be to compute $\Delta_T(\theta)$ at the
current parameter settings $\theta$, and then modify $\theta$ by
$\theta\leftarrow \theta + \gamma
\Delta_T(\theta)$. Provided $\Delta_T(\theta)$ is close enough to the
true gradient direction $\nabla\eta(\theta)$, and provided the
step-sizes $\gamma$ are suitably decreasing, standard stochastic
optimization theory tells us that this technique will converge to
a local maximum of $\eta(\theta)$. However, given that each
computation of $\Delta_T(\theta)$ requires many iterations of the
\pomdp\ to guarantee suitably accurate gradient estimates (that is,
in general $T$ needs to be large), we would like to more
aggressively exploit the information contained in $\Delta_T(\theta)$ than by
simply adjusting the parameters $\theta$ by a small amount in the
direction $\Delta_T(\theta)$.

There are two techniques for making better use of gradient information
that are widely used in {\em non-stochastic} optimization: better
choice of the search direction and better choice of step size. Better
search directions can be found by employing {\em conjugate-gradient}
directions rather than the pure gradient direction. Better step sizes
are usually obtained by performing some kind of line-search to find a
local maximum in the search direction, or through the use of second order
methods. Since line-search techniques tend to be more robust to departures
from quadraticity in the optimization surface, we will only consider those
here \cite<however, see>[Section 7.3, for a discussion of how second-order
derivatives may be computed with a \gpomdp-like algorithm]{jair_01a}.

\conjgrad, described in Algorithm~\ref{algorithm:conjgrad}, is a
version of the Polak-Ribiere conjugate-gradient algorithm \cite<see,
\eg>[Section 5.5.2]{fine99} that is designed to operate using only
noisy (and possibly) biased estimates of the gradient of the objective
function (for example, the estimates $\Delta_T$ provided by \pomdpg).
The argument \grad\ to \conjgrad\ computes the gradient estimate. The
novel feature of \conjgrad\ is \gsearch, a linesearch subroutine that
uses only gradient information to find the local maximum in the search
direction. The use of gradient information ensures \gsearch\ is robust
to noise in the performance estimates. Both \conjgrad\ and \gsearch\
can be applied to any stochastic optimization problem for which noisy
(and possibly) biased gradient estimates are available.

The argument $s_0$ to \conjgrad\ provides an initial step-size for
\gsearch.  The argument $\epsilon$ provides a stopping condition; when
$\|\grad(\theta)\|^2$ falls below $\epsilon$, \conjgrad\ terminates.

\begin{algorithm}
\caption{$\quad\conjgrad(\grad, \theta, s_0, \epsilon)$}
\label{algorithm:conjgrad}
\begin{algorithmic}[1]
\STATE {\bf Given: }
\begin{itemize}
\item $\grad\colon \R^K\to \R^K$: a (possibly noisy and biased)
estimate of the gradient of the objective function to be maximized. 
\item Starting parameters $\theta \in \R^K$ (set to maximum on return).
\item Initial step size $s_0 > 0$.
\item Gradient resolution $\epsilon$. 
\end{itemize}
\STATE $g = h = \grad(\theta)$
\WHILE{$\|g\|^2 \ge \epsilon$}
        \STATE $\gsearch(\grad, \theta, h, s_0, \epsilon)$
        \STATE $\Delta = \grad(\theta)$
        \STATE $\gamma = (\Delta - g) \cdot \Delta / \|g\|^2$
        \STATE $h = \Delta + \gamma h$
        \IF{$h \cdot \Delta < 0$}
        \STATE $h = \Delta$
        \ENDIF
        \STATE $g = \Delta$
\ENDWHILE
\end{algorithmic}
\end{algorithm}

\subsection{The $\mathbold{\gsearch}$ algorithm}

The key to the successful operation of \conjgrad\ is the linesearch
algorithm \gsearch\ (Algorithm~\ref{algorithm:linesearch}). 
\gsearch\ uses only gradient information to bracket the maximum in the 
direction $\theta^*$, and then quadratic interpolation to jump to the
maximum. 

\begin{figure}
\begin{center}
\includegraphics[scale=0.8]{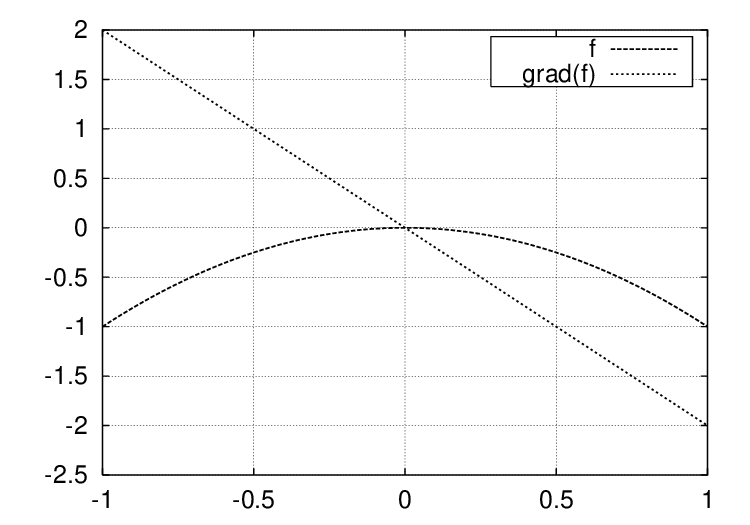}
\caption{Stylized plot of the average reward $\eta(\theta)$ and the
gradient $\nabla\eta(\theta) \cdot \theta^*$ in a search direction
$\theta^*$. 
\label{figure:bracket_clean}}
\end{center}
\end{figure}

\begin{figure}
\begin{center}
\includegraphics[scale=0.8]{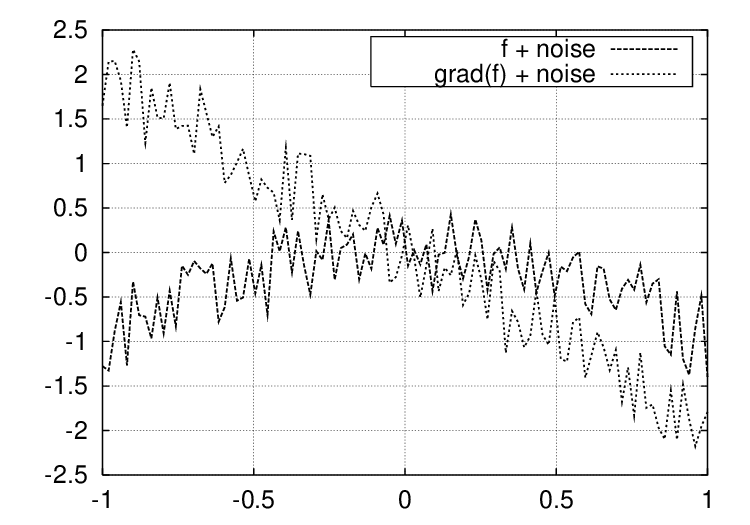}
\caption{Plot as in Figure \ref{figure:bracket_clean} but with
estimation noise added to both the function and gradient curves. 
\label{figure:bracket_noisy}}
\end{center}
\end{figure}

\begin{figure}
\begin{center}
\includegraphics[scale=0.8]{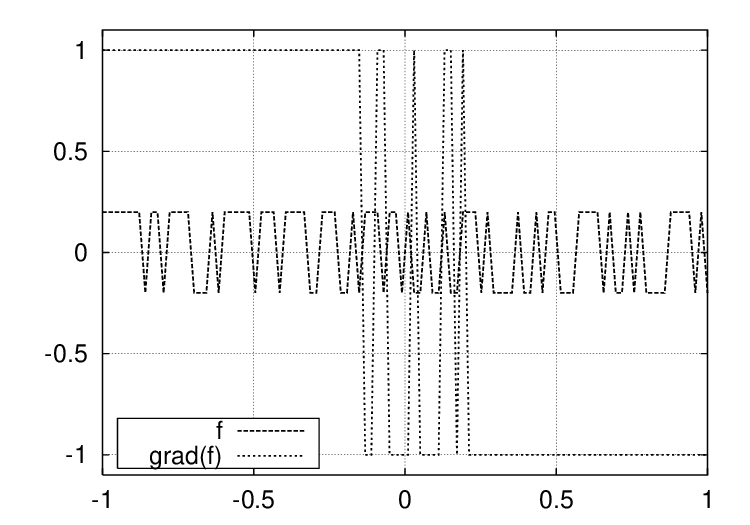}
\caption{Plot of the possible maximum locations that would be found by a
line-search algorithm based on value estimates ($f$), and one based on
gradient estimates (grad($f$)), for the curves in Figure
\ref{figure:bracket_noisy}. The zero-crossings in each case are the
possible locations. Note that the gradient-based approach more
accurately localizes the maximum.
\label{figure:bracket_zero}}
\end{center}
\end{figure}

We found the use of gradients to bracket the maximum far more robust
than the use of function values. To illustrate why this is so, in
Figure~\ref{figure:bracket_clean} we have plotted a stylized view of the
average reward $\eta(\theta)$ along some search direction $\theta^*$
(labeled ``$f$'' in the figure),
and its gradient in that direction $\nabla\eta(\theta)
\cdot\theta^*$ (labeled ``grad($f$)''). There are two ways we could search in the direction
$\theta^*$ to bracket the maximum of $\eta(\theta)$ in that direction
(at $0$ in this case), one using function values and the other using
gradient estimates:
\begin{enumerate} 
\item 
Find three points $\theta_1, \theta_2, \theta_3$,
all lying in the direction $\theta^*$ from $\theta$, such that
$\eta(\theta_1) <\eta(\theta_2)$ and $\eta(\theta_3) <
\eta(\theta_2)$. Assuming no overshooting, we then know the maximum
must lie between $\theta_1$ and $\theta_3$ and we can use the three
points and quadratic interpolation to estimate the location of the
maximum. 
\item
Find two points $\theta_1$ and $\theta_2$ such that $\nabla\eta(\theta_1)
\cdot\theta^* > 0$ and $\nabla\eta(\theta_2)\cdot\theta^* < 0$, and again
use quadratic interpolation (which corresponds to linear interpolation
of the gradients) to estimate the location of the maximum. 
\end{enumerate}
Both of these approaches will be equally satisfactory provided there
is no noise in either the function estimates $\eta(\theta)$, or the
gradient estimates $\nabla\eta(\theta)$.  However, when estimates of
$\eta(\theta)$ or $\nabla\eta(\theta)$ are available only through
simulation, they will necessarily be noisy and the situation will look
more like Figure \ref{figure:bracket_noisy}. In this case the use of
gradients to bracket the maximum becomes more desirable, because the
line-search technique based on value estimates could choose any of the
peaks in the plot of $f + \text{noise}$ as the location of the
maximum, which occur nearly uniformly along the $x$-axis, whereas the
second technique based on gradients would choose any of the {\em
  zero-crossings} of the noisy gradient plot, which are far closer to
the true maximum\footnote{There is an implicit assumption in our
  argument that the noise processes in the gradient and value
  estimates are of approximately the same magnitude. If the variance
  of the value estimates is considerably smaller than the variance of
  the gradient estimates then we would expect bracketing with values
  to be superior. In all our experiments we found gradient bracketing
  to be superior.}. This is illustrated in Figure
\ref{figure:bracket_zero}. 


Another view of this phenomenon is that regardless of the variance of
our estimates of $\eta(\theta)$, the variance of
$\sign\left[\eta(\theta_1) - \eta(\theta_2)\right]$ approaches $1$
(the maximum possible) as $\theta_1$ approaches $\theta_2$. Thus, to
reliably bracket the maximum using noisy estimates of $\eta(\theta)$
we need to be able to reduce the variance of the estimates when
$\theta_1$ and $\theta_2$ are close. In our case this means running
the simulation from which the estimates are derived for longer and
longer periods of time. In contrast, the variance of
$\sign\nabla\eta(\theta_1) \cdot
\theta^*$ (and $\sign\nabla\eta(\theta_2) \cdot \theta^*$) is
independent of the distance between $\theta_1$ and $\theta_2$, and in
particular does not grow as the two points approach one another. 

One disadvantage to using gradient estimates to bracket is that it is
not possible to detect extreme overshooting of the maximum. However,
this can be avoided by using value estimates as a ``sanity check'' to
determine if the value has dropped dramatically, and suitably
adjusting the search if this occurs. 

In Algorithm~\ref{algorithm:linesearch}, lines 5--25 bracket the maximum
by finding a parameter setting $\theta_- = \theta_0 + s_- \theta^*$ such
that $\grad(\theta_-) \cdot \theta^* > -\epsilon$, and a second parameter
setting $\theta_+ = \theta_0 + s_+ \theta^*$ such that $\grad(\theta_+)
\cdot \theta^* < \epsilon$. The reason for $\epsilon$ rather than $0$
in these expressions is to provide some robustness against errors in the
estimates $\grad(\theta)$. It also prevents the algorithm ``stepping to
$\infty$'' if there is no local maximum in the direction $\theta^*$. Note
that we use the same $\epsilon$ as used in \conjgrad\ to determine when
to terminate due to small gradient (line 4 in \conjgrad).

Provided that the signs of the gradients at the bracketing points
$\theta_-$ and $\theta_+$ show that the maximum of the quadratic defined
by these points lies between them, line 27 will jump to the maximum.
Otherwise the algorithm simply jumps to the midpoint between $\theta_-$
and $\theta_+$.

\begin{algorithm}
\caption{$\quad\gsearch(\grad, \theta_0, \theta^*, s_0, \epsilon)$}
\label{algorithm:linesearch}
\begin{algorithmic}[1]
\STATE {\bf Given: }
\begin{itemize}
\item $\grad\colon \R^K\to \R^K$: a (possibly noisy and biased)
estimate of the gradient of the objective function. 
\item Starting parameters $\theta_0 \in \R^K$ (set to maximum on return).
\item Search direction $\theta^*\in \R^K$ with $\grad(\theta_0)
\cdot \theta^* > 0$.
\item Initial step size $s_0 > 0$.
\item Inner product resolution $\epsilon >= 0$. 
\end{itemize}
\STATE $s = s_0$
\STATE $\theta = \theta_0 + s \theta^* $
%\COMMENT{Now bracket the maximum}
\STATE $\Delta =  \grad(\theta)$
\IF{$\Delta \cdot \theta^* < 0$}
        \STATE{Step back to bracket the maximum:}
        \REPEAT
                \STATE $s_+ = s$ 
                \STATE $p_+ = \Delta \cdot \theta^*$ 
                \STATE $s = s/2$
                \STATE $\theta = \theta_0 + s \theta^*$
                \STATE $\Delta = \grad(\theta)$
        \UNTIL{$\Delta \cdot \theta^* > -\epsilon$}
        \STATE $s_- = s$
        \STATE $p_- = \Delta \cdot \theta^*$
\ELSE
        \STATE{Step forward to bracket the maximum:}
        \REPEAT
                \STATE $s_- = s$
                \STATE $p_- = \Delta \cdot \theta^*$
                \STATE $s = 2 s$
                \STATE $\theta = \theta_0 + s \theta^*$
                \STATE $\Delta = \grad(\theta)$
        \UNTIL{$\Delta \cdot \theta^* < \epsilon$}
        \STATE $s_+ = s$
        \STATE $p_+ = \Delta \cdot \theta^*$
\ENDIF
\IF{$p_- > 0$ and $p_+<0$}
        \STATE $s = s_- - p_- \frac{s_+ - s_-}{p_+ - p_-}$
\ELSE 
        \STATE $s = \frac{s_- + s_+}{2}$
\ENDIF
\STATE $\theta_0 = \theta_0 + s \theta^*$
\end{algorithmic}
\end{algorithm}

\subsection{On-line optimization of the average reward: $\mathbold{\olpomdp}$}

\conjgrad\ combined with \gsearch\ operates by iteratively choosing ``uphill'' directions and
then searching for a local maximum in the chosen direction. If the
\grad\ argument to \conjgrad\ is \pomdpg, the optimization will
involve many iterations of the underlying \pomdp\ between parameter
updates.

In traditional stochastic optimization one typically uses algorithms
that update the parameters at {\em every} iteration, rather than
accumulating gradient estimates over many iterations. 
Algorithm~\ref{algorithm:olpgradmdp}, \olpomdp, presents an adaptation
of \gpomdp\ to this form. See \citeA{cdc_00} for a proof that
\olpomdp\ converges to the vicinity of a local maximum of $\eta(\theta)$. Note that
\olpomdp\ is very similar to the algorithms proposed in \citeA{kimura95,kimura97}.

\begin{algorithm}
\caption{$\olpomdp(\beta, T, \theta_0) \rightarrow \R^K$.}
\label{algorithm:olpgradmdp}
\begin{algorithmic}[1]
\STATE {\bf Given: }
\begin{itemize}
\item $\beta \in [0,1)$.
\item $T >0$.
\item Initial parameter values $\theta_0\in \R^K$.
\item Randomized parameterized policies
$\left\{\mu(\theta, \cdot)\colon \theta\in\R^K\right\}$
satisfying Assumption \ref{ass:deriv}. 
\item \pomdp\ with rewards satisfying Assumption~\ref{ass:rbound}, and 
which when controlled by $\mu(\theta, \cdot)$ generates stochastic
matrices $P(\theta)$ satisfying Assumption~\ref{ass:station}.
\item Step sizes $\gamma_t, t=0,1, \dots$ satisfying $\sum \gamma_t =
\infty$ and $\sum\gamma_t^2 < \infty$.  
\item Arbitrary (unknown) starting state $X_0$.
\end{itemize}
\STATE Set $z_0 = 0$ ($z_0 \in\R^K$).
\FOR{$t=0$ to $T-1$}
\STATE Observe $Y_t$ (generated according to $\nu(X_t)$).
\STATE Generate control $U_t$ according to $\mu(\theta, Y_t)$
\STATE Observe $r(X_{t+1})$ (where the next state $X_{t+1}$ is
generated according to $p_{X_tX_{t+1}}(U_t)$. 
\STATE Set $z_{t+1} = \beta z_t +
  \dfrac{\nabla \mu_{U_t}(\theta, Y_t)}{\mu_{U_t}(\theta, Y_t)}$
\STATE Set $\theta_{t+1} = \theta_t + \gamma_t r(X_{t+1}) z_{t+1}$
\ENDFOR
\STATE return $\theta_T$
\end{algorithmic}
\end{algorithm}

\section{Experiments}
\label{sec:exp}
In this section we present several sets of experimental
results. Throughout this section, where we refer to \conjgrad\ we
mean \conjgrad\ with \pomdpg\ as its \grad\ argument. 

In the first set of experiments, we consider a system in which a
controller is used to select actions for a 3-state Markov Decision
Process (\mdp). For this system we are able to compute the true
gradient exactly using the matrix equation
\begin{equation}
\label{eq:gradeta}
\nabla \eta(\theta) = \pi'(\theta) \nabla P(\theta)\left[I - P(\theta) + e
\pi'(\theta)\right]^{-1} r,
\end{equation}
where $P(\theta)$ is the transition matrix of the underlying Markov
chain with the controller's parameters set to $\theta$, $\pi'(\theta)$
is the stationary distribution corresponding to $P(\theta)$ (written
as a row vector), $e\pi'(\theta)$ is the square matrix in which each row is
the stationary distribution, and $r$ is the (column) vector of rewards
%PB I think they're the same now??
\cite<see >[Section 3, for a derivation of \eqref{eq:gradeta}]{jair_01a}.
Hence we can compare the estimates $\Delta_T$
generated by \pomdpg\ with the true gradient
$\nabla\eta(\theta)$, both as a function of the number of iterations $T$ and
as a function of the discount parameter $\beta$. We also optimize the 
performance of the controller using the
on-line algorithm, \olpomdp, and the off-line algorithm 
\conjgrad. \conjgrad\ 
reliably converges to a near optimal policy with around 100 iterations
of the \mdp, while the on-line method requires approximately 1000
iterations.  This should be contrasted with training a linear {\em
value-function} for this system using $\TD(1)$
\cite{sutton88}, which can be shown to converge to a value function whose
one-step lookahead policy is suboptimal \cite{tr_std_99}.

In the second set of experiments, we consider a simple ``puck-world''
problem in which a small puck must be navigated around a
two-dimensional world by applying thrust in the $x$ and $y$
directions. We train a 1-hidden-layer neural-network controller for
the puck using \conjgrad. Again the controller reliably converges to
near optimality. 

In the third set of experiments we use \conjgrad\ to
optimize the admission thresholds for the call-admission problem
considered in
\cite{marbachthesis98}. 

In the final set of experiments we use \conjgrad\ to train a switched neural-network
controller for a two-dimensional variant of the ``mountain-car'' task 
\cite[Example 8.2]{sutton98}.
 
In all the experiments we found that convergence of the line-searches
was greatly improved if all calls to the \gpomdp\ algorithm were
seeded with the same random number sequence.

\subsection{A three-state MDP}
\label{subsec:3state}

In this section we consider a three-state \mdp, in each state
of which there is a choice of two actions $a_1$ and $a_2$.
Table~\ref{tab:threestate} shows the transition probabilities as a
function of the states and actions. Each state $x$ has an associated
two-dimensional feature vector $\phi(x) = (\phi_1(x), \phi_2(x))$ and
reward $r(x)$ which are detailed in  Table \ref{tab:values}. Clearly,
the optimal policy is to always select the action that leads to state
$C$ with the highest probability, which from Table \ref{tab:threestate}
means always selecting action $a_2$.

\begin{table}
\begin{center}
\begin{tabular}{|c|c|c|c|c|}
\hline
Origin &  &
\multicolumn{3}{|c|}{Destination State Probabilities}\\
State & Action  & $A$    & $B$    &    $C$ \\
\hline
$A$       &   $a_1$   & ~~~~~0.0~~~~  & ~~~~0.8~~~~ & 0.2 \\
$A$       &   $a_2$   & ~~~~~0.0~~~~  & ~~~~0.2~~~~ & 0.8 \\ \hline
$B$       &   $a_1$   & ~~~~~0.8~~~~  & ~~~~0.0~~~~ & 0.2 \\
$B$       &   $a_2$   & ~~~~~0.2~~~~  & ~~~~0.0~~~~ & 0.8 \\ \hline
$C$       &   $a_1$   & ~~~~~0.0~~~~  & ~~~~0.8~~~~ & 0.2  \\
$C$       &   $a_2$   & ~~~~~0.0~~~~  & ~~~~0.2~~~~ & 0.8 \\
\hline
\end{tabular}
\caption{Transition probabilities of the three-state MDP\label{tab:threestate}}
\end{center}
\end{table}

\begin{table}
\begin{center}
\begin{tabular}{lcccc}
  $r(A)=0$ & & $\phi_1(A)=\frac{12}{18}$  & $\phi_2(A)=\frac{6}{18}$\\[2mm]
  $r(B)=0$ & & $\phi_1(B)=\frac{6}{18}$ & $\phi_2(B)=\frac{12}{18}$\\[2mm]
  $r(C)=1$ & & $\phi_1(C) = \frac{5}{18}$ & $\phi_2(C) = \frac{5}{18}$
\end{tabular}
\caption{Three-state rewards and features.\label{tab:values}}
\end{center}
\end{table}

This rather odd choice of feature vectors for the states ensures that
a value function linear in those features and trained using
$\TD(1)$---while observing the optimal policy---will implement a
suboptimal greedy one-step lookahead policy (see
\cite{tr_std_99} for a proof). Thus, in contrast to the gradient based approach, 
for this system, $\TD(1)$ training a linear value function is
guaranteed to produce a worse policy if it
starts out observing the optimal policy.

\subsubsection{Training a controller}
Our goal is to learn a stochastic controller for this system
that implements an optimal (or near-optimal) policy. Given a parameter
vector $\theta = (\theta_1,\theta_2, \theta_3, \theta_4)$, we generate
a policy as follows. For any state $x$, let
\begin{align*}
s_1(x) &:= \theta_1 \phi_1(x) + \theta_2 \phi_2(x) \\
s_2(x) &:= \theta_3 \phi_1(x) + \theta_4 \phi_2(x). 
\end{align*}
Then the probability of choosing action $a_1$ in state $x$ is given by 
$$
\mu_{a_1}(x) = \frac{e^{s_1(x)}}{ e^{s_1(x)} + e^{s_2(x)}},
$$
while the probability of choosing action $a_2$ is given by
$$
\mu_{a_2}(x) = \frac{e^{s_2(x)}}{ e^{s_1(x)} + e^{s_2(x)}} = 1 - \mu_{a_1}(x). 
$$
The ratios $\frac{\nabla \mu_{a_i}(x)}{\mu_{a_i}(x)}$ needed by
Algorithms \ref{algorithm:pgradmdp} and \ref{algorithm:olpgradmdp} are
given by,
\begin{align}
\frac{\nabla \mu_{a_1}(x)}{\mu_{a_1}(x)}  &= 
\frac{e^{s_2(x)}}{e^{s_1(x)} + e^{s_2(x)}}\left[\phi_1(x), \phi_2(x),
-\phi_1(x), -\phi_2(x)\right] \\
\frac{\nabla \mu_{a_2}(x)}{\mu_{a_2}(x)}  &= 
\frac{e^{s_1(x)}}{e^{s_1(x)} + e^{s_2(x)}}\left[-\phi_1(x), -\phi_2(x),
\phi_1(x), \phi_2(x)\right] 
\end{align}
Since the second two components in $\nabla\mu/\mu$ are always the
negative of the first two, this shows that two of the parameters are
redundant in this case: we could just as well have set $\theta_3 = -
\theta_1$ and $\theta_4 = -\theta_2$. 

\subsubsection{Gradient estimates}

\begin{figure}
\begin{center}
\includegraphics[scale=0.45]{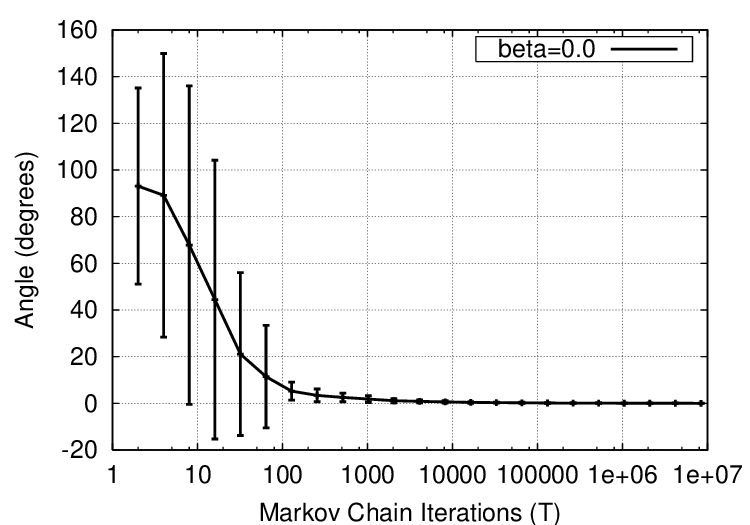}
\includegraphics[scale=0.45]{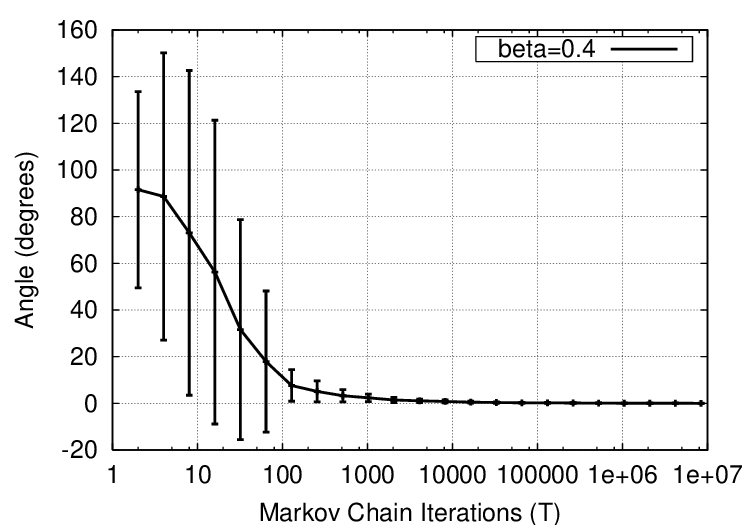}
\includegraphics[scale=0.45]{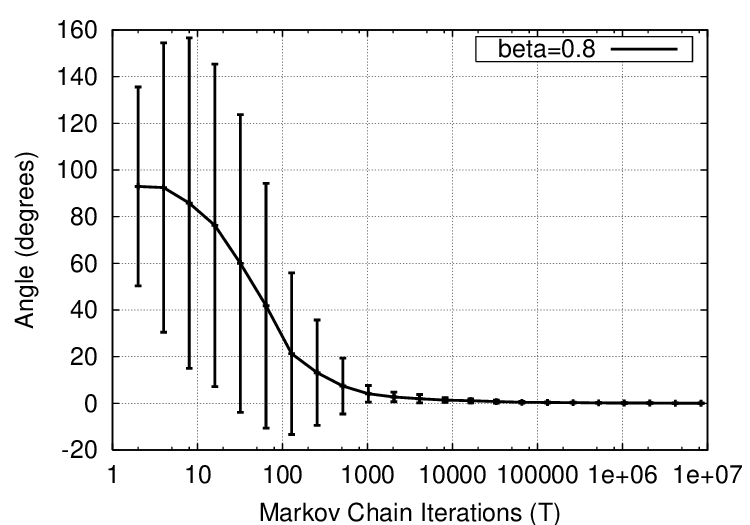}
\includegraphics[scale=0.45]{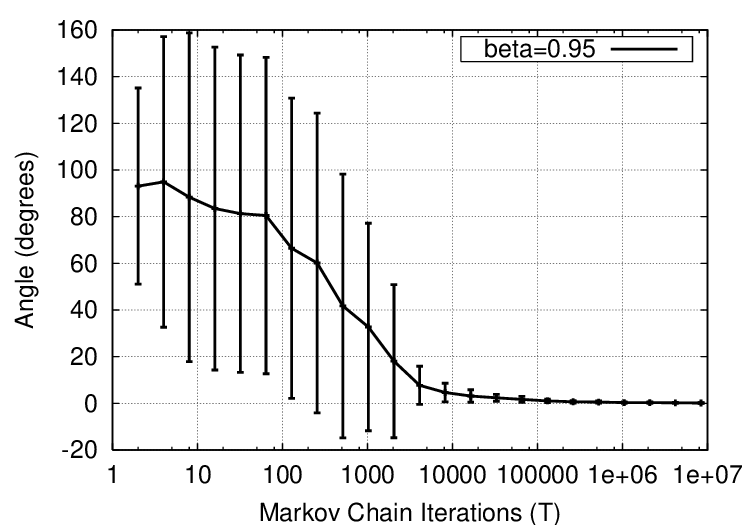}
\caption{Angle between the true gradient $\nabla \eta$ and the
estimate $\Delta_T$ for the three-state Markov chain, for various
values of the discount parameter $\beta$. $\Delta_T$ was generated by
Algorithm \ref{algorithm:pgradmdp}. Averaged over 500 independent
runs. Note the higher variance at large $T$ for the larger values of
$\beta$. Error bars are one standard deviation.\label{fig:3stateanglesvar}}
\end{center}
\end{figure}

\begin{figure}
\begin{center}
\includegraphics[scale=0.45]{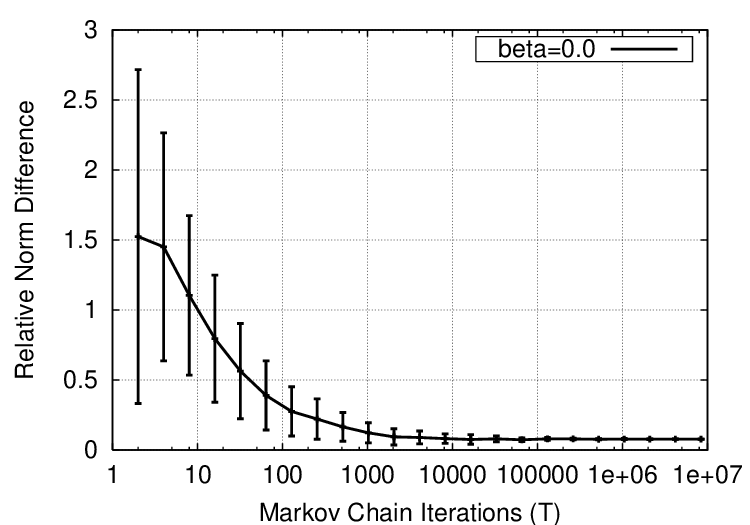}
\includegraphics[scale=0.45]{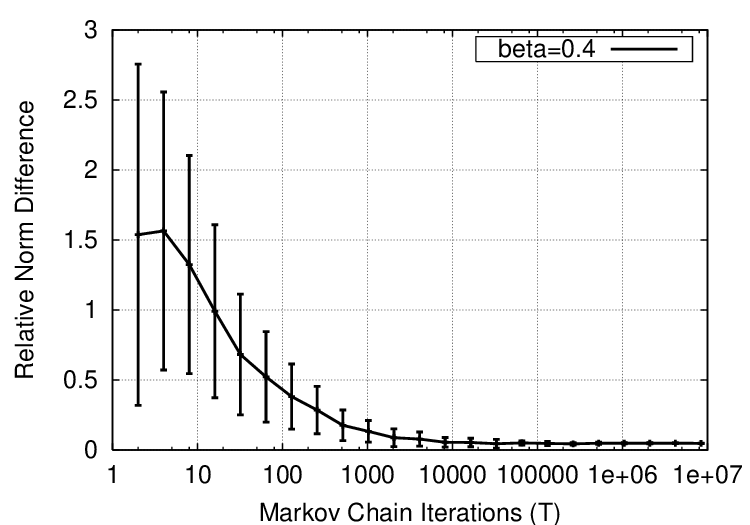}
\includegraphics[scale=0.45]{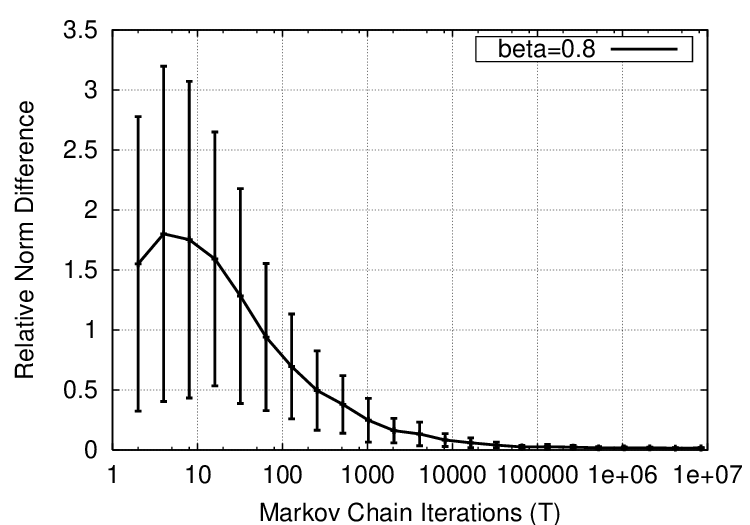}
\includegraphics[scale=0.45]{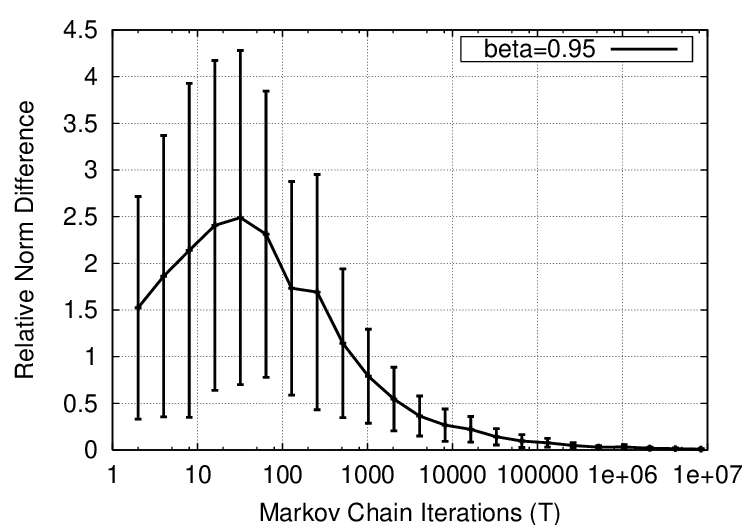}
\caption{A plot of $\frac{\|\nabla \eta - \Delta_T\|}{\|\nabla \eta\|}$
for the three-state Markov chain, for various values of the discount
parameter $\beta$. $\Delta_T$ was generated by Algorithm
\ref{algorithm:pgradmdp}. Averaged over 500 independent runs. Note the
higher variance at large $T$ for the larger values of
$\beta$. Error bars are one standard deviation.\label{fig:3statenormsvar}}
\end{center}
\end{figure}

\begin{figure}
\begin{center}
\includegraphics[scale=0.8]{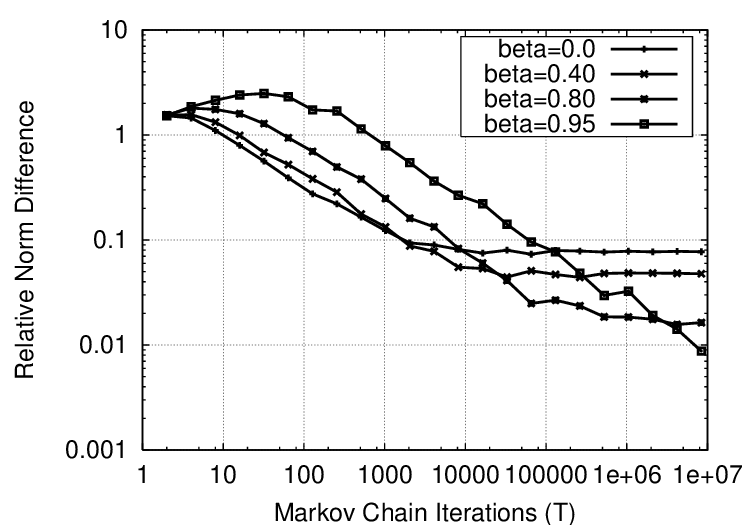}
\caption{Graph showing the error in the
estimate $\Delta_T$ (as measured by $\frac{\|\nabla \eta -
\Delta_T\|}{\|\nabla \eta\|}$) for various values of $\beta$ for the
three-state Markov chain. $\Delta_T$ was generated by Algorithm
\ref{algorithm:pgradmdp}. Note the decrease in the final bias as
$\beta$ increases. Both axes are log scales.
\label{fig:3statenormsbias}}
\end{center}
\end{figure}

With a parameter vector\footnote{Other initial values of the parameter
vector were chosen with similar results. Note that $[1,1,-1,-1]$ generates
a suboptimal policy.}
of $\theta = \left[1, 1, -1, -1\right]$,
\pomdpg\ was used to generate estimates $\Delta_T$ of
$\nb \eta$, for various values of $T$
and $\beta\in [0,1)$. To measure the progress of $\Delta_T$ towards
the true gradient $\nabla \eta$, $\nabla \eta$ was calculated from
\eqref{eq:gradeta} and then for each value of $T$ the {\em angle} between
$\Delta_T$ and $\nabla \eta$ and the relative error $\frac{\|\Delta_T
- \nabla \eta\|}{\|\nabla\eta\|}$ were recorded.  The angles and
relative errors are plotted in Figures~\ref{fig:3stateanglesvar},
\ref{fig:3statenormsvar} and~\ref{fig:3statenormsbias}.

The graphs illustrate a typical trade-off for the \pomdpg\ algorithm:
small values of $\beta$ give higher bias in the estimates, while
larger values of $\beta$ give higher variance (the final bias is only shown
in Figure~\ref{fig:3statenormsbias} for the norm deviation because
it was too small to measure for the angular deviation). The
bias introduced by having $\beta < 1$ is very small for this system. In
the worst case, $\beta = 0.0$, the final gradient {\em direction} is
indistinguishable from the true direction while the relative deviation
$\frac{\|\nabla \eta - \Delta_T\|}{\|\nabla \eta\|}$ is only $7.7\%$.

\subsubsection{Training via conjugate-gradient ascent}
\label{mdpcg}
\sloppy
\conjgrad\ with \pomdpg\ as the ``\grad'' argument was used to train
the parameters of the controller described in the previous
section. Following the low bias observed in the experiments of the
previous section, the argument $\beta$ of \pomdpg\ was set to
$0$. After a small amount of experimentation, the arguments $s_0$ and
$\epsilon$ of \conjgrad\ were set to $100$ and $0.0001$
respectively. None of these values were critical, although the
extremely large initial step-size ($s_0$) did considerably reduce the
time required for the controller to converge to near-optimality.

We tested the performance of \conjgrad\ for a range of values of the
argument $T$ to \pomdpg\ from $1$ to $4096$. Since \gsearch\ only uses
\pomdpg\ to determine the {\em sign} of the inner product of the
gradient with the search direction, it does not need to run \pomdpg\
for as many iterations as \conjgrad\ does. Thus, \gsearch\ determined
its own $T$ parameter to \pomdpg\ as follows. Initially, (somewhat
arbitrarily) the value of $T$ within \gsearch\ was set to $1/10$ the
value used in \conjgrad\ (or 1 if the value in \conjgrad\ was less
than 10). \gsearch\ then called \pomdpg\ to obtain an estimate
$\Delta_T$ of the gradient direction. If $\Delta_T \cdot \theta^* < 0$
($\theta^*$ being the desired search direction) then $T$ was doubled
and \gsearch\ was called again to generate a new estimate
$\Delta_T$. This procedure was repeated until $\Delta_T \cdot \theta^*
> 0$, or $T$ had been doubled four times. If $\Delta_T \cdot \theta^*$
was still negative at the end of this process, \gsearch\ searched for
a local maximum in the direction $-\theta^*$, and the number of
iterations $T$ used by \conjgrad\ was doubled on the next iteration
(the conclusion being that the direction $\theta^*$ was generated by
overly noisy estimates from \pomdpg).

\begin{figure}
\begin{center}
\includegraphics[scale=0.8]{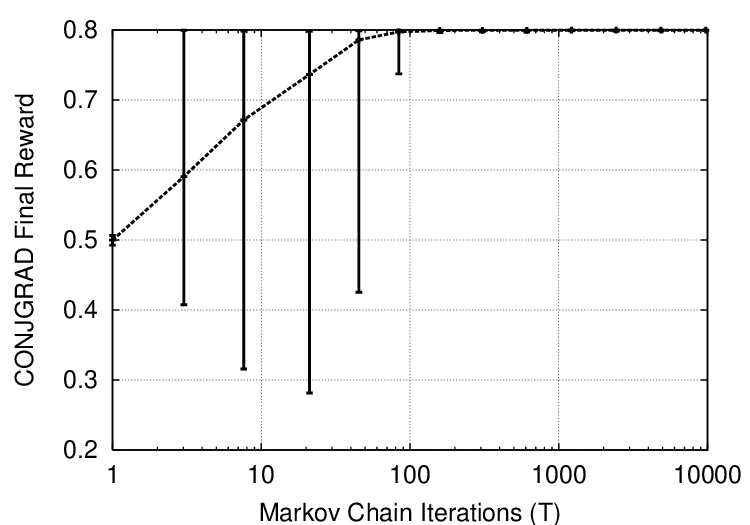}
\caption{Performance of the 3-state Markov chain controller trained
by \conjgrad\ as a function of the total number of iterations of the
Markov chain. The performance was computed exactly from the stationary
distribution induced by the controller. The average reward of the optimal
policy is $0.8$. Averaged over 500 independent runs. The error bars
were computed by dividing the results into two separate bins depending
on whether they were above or below the mean, and then computing the
standard deviation within each bin.
\label{fig:3statecg}}
\end{center}
\end{figure}

Figure~\ref{fig:3statecg} shows the average reward $\eta(\theta)$ of
the final controller produced by \conjgrad, as a function of the total
number of simulation steps of the underlying Markov chain.  The plots
represent an average over $500$ independent runs of \conjgrad. Note
that $0.8$ is the average reward of the optimal policy. The parameters
of the controller were (uniformly) randomly initialized in the range
$[-0.1, 0.1]$ before each call to \conjgrad. After each call to
\conjgrad, the average reward of the resulting controller was computed
exactly by calculating the stationary distribution for the
controller. From Figure~\ref{fig:3statecg}, optimality is reliably
achieved using approximately 100 iterations of the Markov chain.

\subsubsection{Training on-line with \olpomdp}
\begin{figure}
\begin{center}
\includegraphics[scale=0.45]{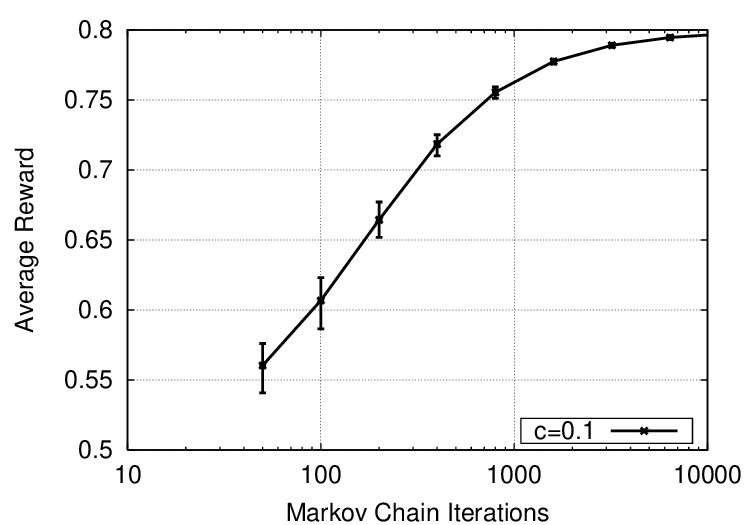}
\includegraphics[scale=0.45]{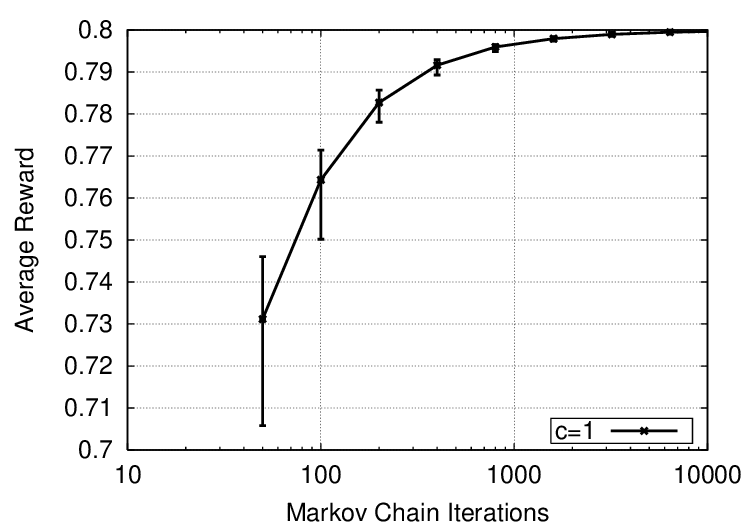}
\includegraphics[scale=0.45]{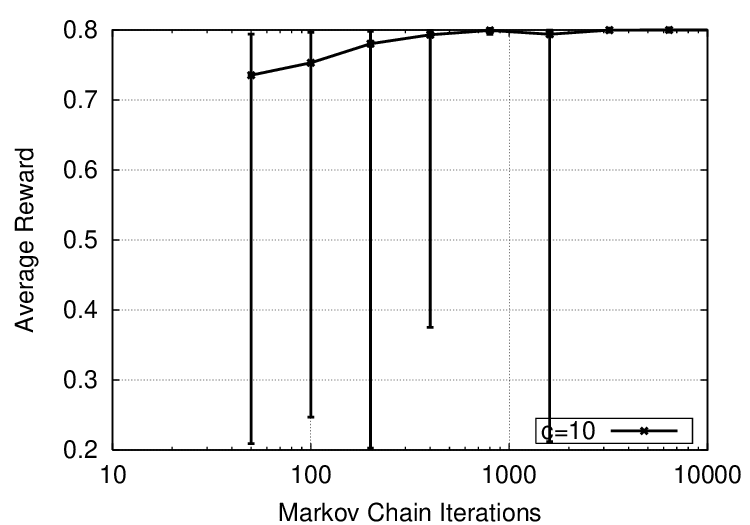}
\includegraphics[scale=0.45]{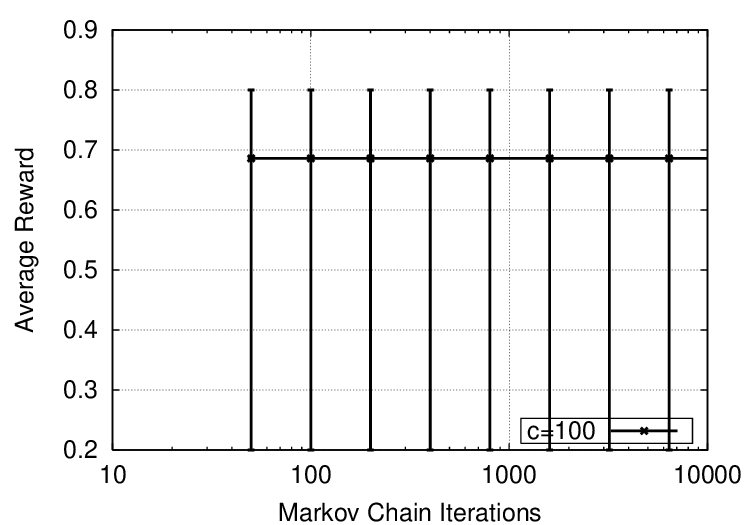}
\caption{Performance of the 3-state Markov chain controller as a function of the number of
iteration steps in the {\em on-line} algorithm, Algorithm
\ref{algorithm:olpgradmdp}, for fixed step sizes of $0.1, 1, 10$, and
$100$. Error bars were computed as in Figure \ref{fig:3statecg}. \label{fig:chainol}}
\end{center}
\end{figure}

The controller was also trained on-line using Algorithm
\ref{algorithm:olpgradmdp} (\olpomdp) with fixed step-sizes $\gamma_t
= c$ with $c=0.1, 1, 10, 100$. Reducing step-sizes of the form
$\gamma_t = c/t$ were tried, but caused intolerably slow convergence.
Figure~\ref{fig:chainol} shows the performance of the controller
(measured exactly as in the previous section) as a function of the
total number of iterations of the Markov chain, for different values
of the step-size $c$. The graphs are averages over 100 runs, with the
controller's weights randomly initialized in the range $[-0.1, 0.1]$
at the start of each run. From the figure, convergence to optimal is
about an order of magnitude slower than that achieved by \conjgrad, for
the best  step-size of $c=1.0$. Step-sizes much greater that $c=10.0$
failed to reliably converge to an optimal policy.

\subsection{Puck World}
\label{subsec:puck}

In this section, experiments are described in which \conjgrad\ and
\olpomdp\ were used to train 1-hidden-layer neural-network controllers
to navigate a small puck around a two-dimensional world.

\subsubsection{The World}
The puck was a unit-radius, unit-mass disk
constrained to move in the plane in a region 100 units square. The
puck had no internal dynamics (i.e rotation). Collisions with the
region's boundaries  were inelastic with a (tunable) coefficient of restitution $e$
(set to $0.9$ for the experiments reported here). The puck was
controlled by applying a 5 unit force in either the positive or
negative $x$ direction, and a 5 unit force in either the positive or
negative $y$ direction, giving four different controls in total. The
control could be changed every $1/10$ of a second, and the simulator
operated at a granularity of $1/100$ of a second. The puck also had a
retarding force due to air resistance of $0.005 \times
\text{speed}^2$. There was no friction between the puck and the
ground.

The puck was given a reward at each decision point ($1/10$ of a
second) equal to $-d$ where $d$ was the distance between the puck and
some designated target point.  To encourage the controller to learn to
navigate the puck to the target independently of the starting state,
the puck state was reset every 30 (simulated) seconds to a random
location and random $x$ and $y$ velocities in the range $[-10, 10]$,
and at the same time the target position was set to a random location. 

Note that the size of the state-space in this example is essentially
infinite, being of the order of $2^{\text{PRECISION}}$ where
$\text{PRECISION}$ is the floating point precision of the machine ($64$
bits). Thus, the time between visits to a recurrent state is likely
to be large. Also, the puck cannot just maximize its immediate reward
because this leads to significant overshooting of the target locations.

\subsubsection{The controller} 
\label{sec:puckcontrol}
A one-hidden-layer neural-network with six input nodes, eight hidden
nodes and four output nodes was used to generate a probabilistic
policy in a similar manner to the controller in the three-state
Markov chain example of the previous section. Four of the inputs were set to the
raw $x$ and $y$ locations and velocities of the puck at the current
time-step, the other two were the differences between the puck's $x$
and $y$ location and the target's $x$ and $y$ location
respectively. The location inputs were scaled to lie between $-1$ and
$1$, while the velocity inputs were scaled so that a speed of $10$
units per second mapped to a value of $1$. 
The hidden nodes computed a $\tanh$ squashing function,
while the output nodes were linear. Each hidden and output node had
the usual additional offset parameter. The four output nodes were
exponentiated and then normalized as in the Markov-chain example to
produce a probability distribution over the four controls ($\pm 5$
units thrust in the $x$ direction, $\pm 5$ units thrust in the $y$
direction). Controls were selected at random from this
distribution.

\subsubsection{Conjugate gradient ascent}
\label{sec:puckgrad}
We trained the neural-network controller using \conjgrad\ with the
gradient estimates generated by \pomdpg.  After some experimentation 
we chose $\beta=0.95$ and $T=1,000,000$ as the
parameters \conjgrad\ supplied to \pomdpg. \gsearch\ used the same value
of $\beta$ and the scheme discussed in Section~\ref{mdpcg} to
determine the number of iterations with which to call \pomdpg. 

Due to the saturating nature of the neural-network hidden nodes
(and the exponentiated output nodes), there was a tendency for the
network weights to converge to local minima at ``infinity''. That is,
the weights would grow very rapidly early on in the simulation, but
towards a suboptimal solution. Large weights tend to imply very small
gradients and thus the network becomes ``stuck'' at these suboptimal
solutions. We have observed a similar behaviour when training neural
networks for pattern classification problems. To fix the problem, we
subtracted a small quadratic penalty term $\gamma \|\theta\|^2$ from
the performance estimates and hence also a small correction $2\gamma
\theta_i$ from the gradient calculation\footnote{When used as a technique
for capacity control in pattern classification, this technique goes by
the name ``weight decay''. Here we used it to condition the optimization
problem.}
for $\theta_i$.

We used a decreasing schedule for the quadratic penalty weight
$\gamma$ (arrived at through some experimentation). $\gamma$ was
initialized to $0.5$ and then on every tenth iteration of \conjgrad,
if the performance had improved by less than 10\% from the value ten
iterations ago, $\gamma$ was reduced by a factor of 10. This schedule
solved nearly all the local minima problems, but at the expense of
slower convergence of the controller.

\begin{figure}
\begin{center}
\includegraphics[scale=0.8]{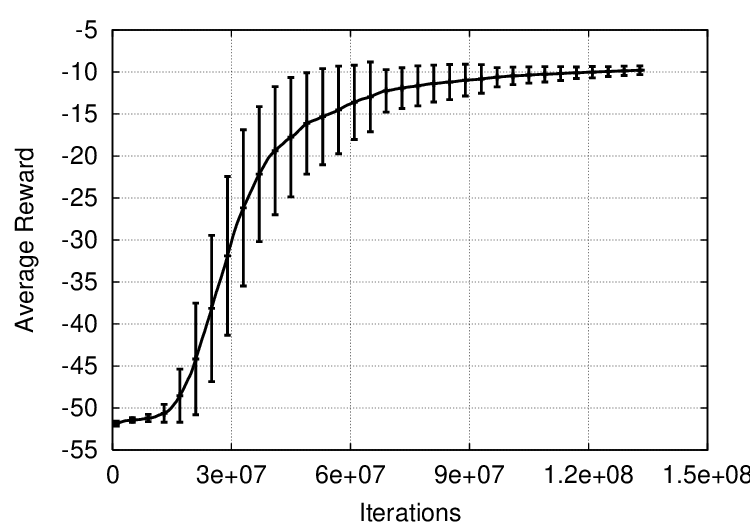}
\caption{Performance of the neural-network puck controller as a
function of the number of iterations of the puck world, when trained
using \conjgrad. Performance estimates were generated by simulating
for $1,000,000$ iterations. Averaged over 100 independent
runs (excluding the four bad runs in Figure \ref{fig:badpuckcg}).\label{fig:puckperform}}
\end{center}
\end{figure}

\begin{figure}
\begin{center}
\includegraphics[scale=0.8]{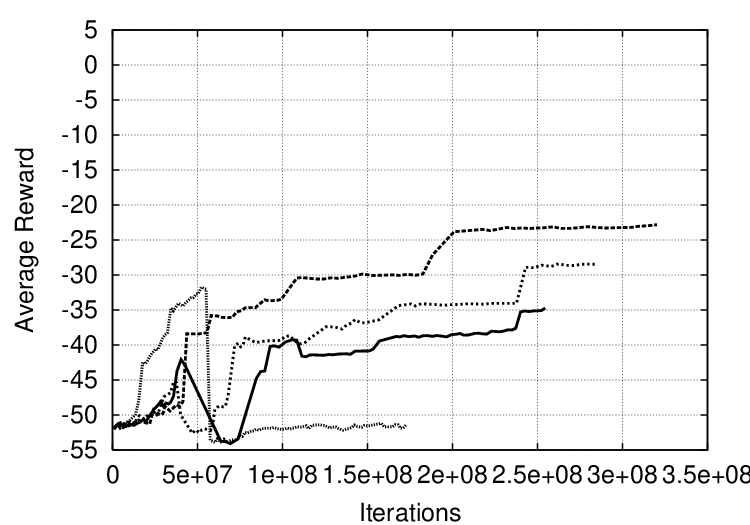}
\caption{Plots of the performance of the neural-network puck
controller for the four runs (out of 100) that converged to
substantially suboptimal local minima. \label{fig:badpuckcg}}
\end{center}
\end{figure}

A plot of the average reward of the neural-network controller is shown in 
Figure \ref{fig:puckperform}, as a function of the number of
iterations of the \pomdp. The graph is an average over 100 independent
runs, with the parameters initialized randomly in the range $[-0.1,0.1]$
at the start of each run. The four bad runs shown in Figure
\ref{fig:badpuckcg} were omitted  from the average because they gave
misleadingly large error bars. 
 
Note that the optimal performance (within the neural-network
controller class) seems to be around $-8$
for this problem, due to the fact that the puck and target locations
are reset every $30$ simulated seconds and hence there is a fixed
fraction of the time that the puck must be away from the target. From
Figure \ref{fig:puckperform} we see that the final performance of the puck
controller is close to optimal. In only 4 of the 100 runs did
\conjgrad\ get stuck in a suboptimal local minimum. Three of those
cases were caused by overshooting in \gsearch\ (see Figure
\ref{fig:badpuckcg}), which could be prevented by adding extra checks
to \conjgrad. 

\begin{figure}
\begin{center}
\includegraphics[scale=0.8]{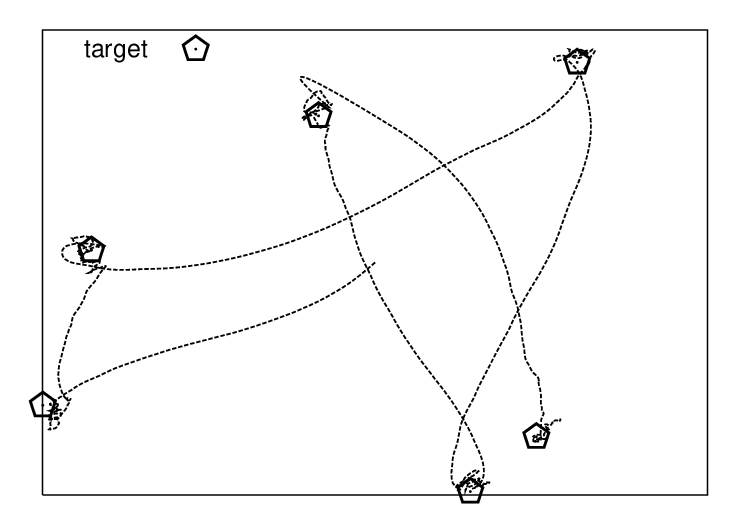}
\caption{Illustration of the behaviour of a typical trained puck
controller. \label{fig:puckbehaviour}}
\end{center}
\end{figure}

Figure \ref{fig:puckbehaviour} illustrates the behaviour of a typical
trained controller. For the purpose of the illustration, only the
target location and puck velocity were randomized every 30 seconds,
not the puck location.

\subsection{Call Admission Control}
\label{sec:call}
\sloppy
In this section we report the results of experiments in which
\conjgrad\ was  applied to
the task of training a controller for the call admission problem
treated by \citeA[Chapter 7]{marbachthesis98}.

\subsubsection{The Problem}
The call admission control problem treated by \citeA[Chapter
7]{marbachthesis98} models the situation in which a telecommunications
provider wishes to sell bandwidth on a communications link to
customers in such a way as to maximize long-term average reward.

Specifically, the problem is a queuing problem. There are three
different types of call, each with its own call arrival rate
$\alpha(1)$, $\alpha(2)$, $\alpha(3)$, bandwidth demand $b(1)$,
$b(2)$, $b(3)$ and average holding time $h(1)$, $h(2)$, $h(3)$. The
arrivals are Poisson distributed while the holding times are
exponentially distributed. The link has a maximum bandwidth of 10
units. When a call arrives and there is sufficient available
bandwidth, the service provider can choose to accept or reject the
call (if there is not enough available bandwidth the call is always
rejected). Upon accepting a call of type $m$, the service provider
receives a reward of $r(m)$ units. The goal of the service provider is
to maximize the long-term average reward. 

The parameters associated
with each call type are listed in Table \ref{table:callparams}. 
With these settings, the optimal policy (found by dynamic
programming by \citeA{marbachthesis98}) is to always accept calls of
type 2 and 3 (assuming sufficient available bandwidth) and to accept
calls of type 1 if the available bandwidth is at least 3. This policy
has an average reward of $0.804$, while the ``always accept'' policy
has an average reward\footnote{There is some discrepancy
between our average rewards and those quoted by
\citeA{marbachthesis98}. This is probably due to a discrepancy in the
way the state transitions are counted, which was not clear from the
discussion in \cite{marbachthesis98}.}
of $0.784$.

\begin{table}
\begin{center}
\begin{tabular}{|lr|r|r|r|}
\hline
Call Type & & 1 & 2 & 3 \\ \hline
Bandwidth Demand & $b$ & 1 & 1 & 1 \\
Arrival Rate & $\alpha$ & $1.8$ & $1.6$ &  $1.4$ \\
Average Holding Time & $h$ & $0.6$ & $0.5$ & $0.4$ \\
Reward & $r$ & 1 & 2 & 4 \\\hline
\end{tabular}
\caption{Parameters of the call admission control problem.\label{table:callparams}} 
\end{center}
\end{table}

\subsubsection{The Controller}
The controller had three parameters $\theta = (\theta_1, \theta_2,
\theta_3)$, one for each type of call. Upon arrival of a call of type
$m$, the controller chooses to accept the call with probability 
$$
\mu(\theta) = \begin{cases} \frac{1}{1 + \exp(1.5(b - \theta_m))}
&\text{ if $b + b(m) \leq 10$,} \\
0 &\text{ otherwise,}
\end{cases}
$$
where $b$ is the currently used bandwidth. This is the class of
controllers studied by \citeA{marbachthesis98}.

\subsubsection{Conjugate gradient ascent}
\begin{figure}
\begin{center}
\includegraphics[scale=0.8]{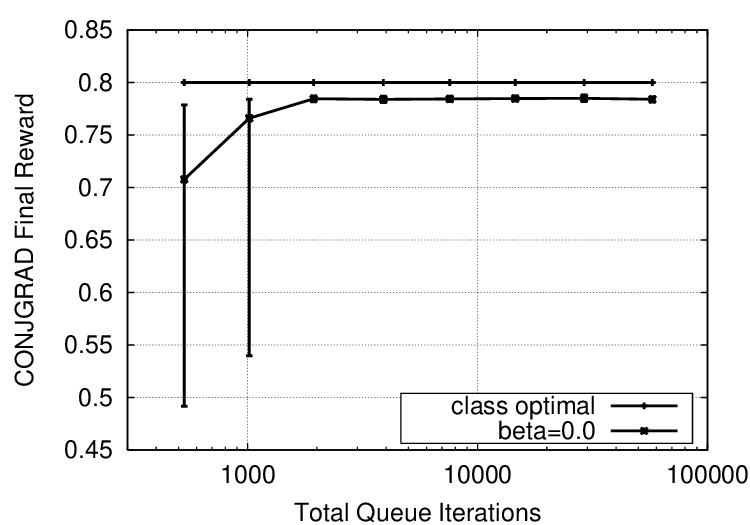}
\caption{Performance of the call admission controller trained by
\conjgrad\, as a function of the total number of iterations of the
queue. The performance was computed by simulating the controller for
100,000 iterations.  The average reward of the globally optimal policy
is $0.804$, the average reward of the optimal policy within the class is
$0.8$, and the plateau performance of \conjgrad\ is $0.784$.  The graphs
are averages from 100 independent runs.
\label{fig:callcg}}
\end{center}
\end{figure}

\conjgrad\ was used to train the above controller, with \pomdpg\
generating the gradient estimates from a range of values of $\beta$
and $T$. The influence of $\beta$ on the performance of the trained
controllers was marginal, so we set $\beta=0.0$ which gave the
lowest-variance estimates. We used the same value of $T$ for calls to
\pomdpg\ within \conjgrad\ and within \gsearch, and this was varied
between $10$ and $10,000$.  The controller was
always started from the same parameter setting $\theta = (8,8,8)$ (as
was done by \citeA{marbachthesis98}). The value of this initial policy
is $0.691$. The graph of the average reward of
the final controller produced by \conjgrad\ as a function of the total
number of iterations of the queue is shown in Figure~\ref{fig:callcg}.
A performance of $0.784$ was reliably achieved with
less than $2000$ iterations of the queue. 

\begin{figure}
\begin{center}
\includegraphics[scale=0.8]{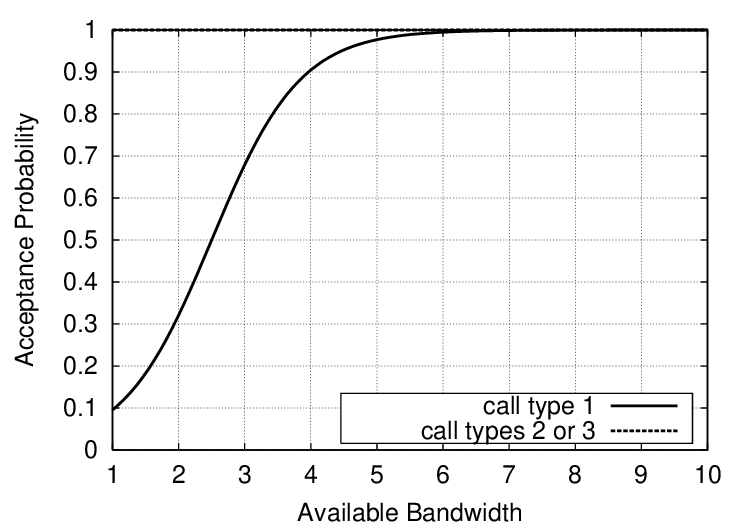}
\caption{Probability of accepting a call of each type under the
call admission policy with near-optimal parameters $\theta_1 = 7.5, \theta_2 =
\theta_3 = 15$. Note that calls of type 2 and 3 are essentially always
accepted. 
\label{fig:accept}}
\end{center}
\end{figure}

Note that the optimal policy is not achievable with this controller class
since it is incapable of implementing any threshold policy other than the
``always accept'' and ``always reject'' policies. Although not provably
optimal, a parameter setting of $\theta_1\approx 7.5$ and any suitably
large values of $\theta_2$ and $\theta_3$ generates something close to
the optimal policy within the controller class, with an average reward
of $0.8$. Figure~\ref{fig:accept} shows the probability of accepting
a call of each type under this policy (with $\theta_2=\theta_3 = 15$),
as a function of the available bandwidth.

\begin{figure}
\begin{center}
\includegraphics[scale=0.8]{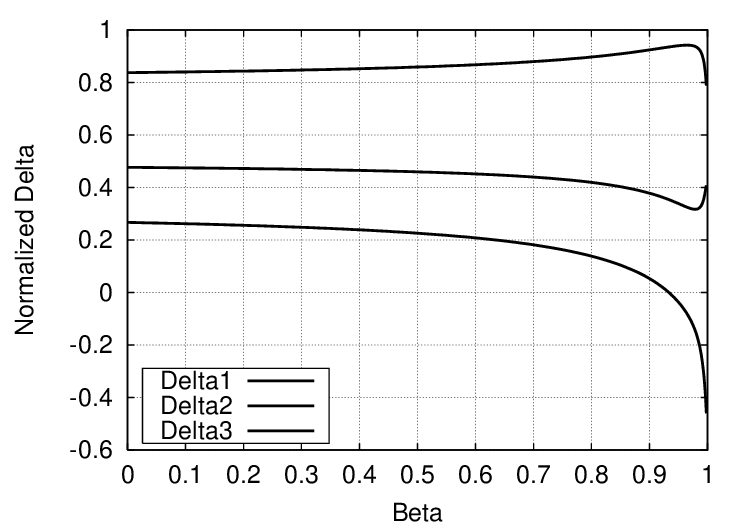}
\caption{Plot of the three components of $\Delta_T$ 
for the call admission problem, as a function of the discount
parameter $\beta$. The parameters were set at $\theta = (8,8,8)$.
$T$ was set to $1,000,000$. Note that $\Delta_{1}$ does not become
negative (the correct sign) until $\beta\approx 0.93$. 
\label{fig:callgrad}}
\end{center}
\end{figure}

The controllers produced by \conjgrad\ with $\beta=0.0$ and sufficiently
large $T$ are essentially ``always accept'' controllers with an average
reward of $0.784$, within 2\% of the optimum achievable in the class. To
produce policies even nearer to the optimal policy in performance,
\conjgrad\ must keep  $\theta_1$ close to its starting value of $8$,
and hence the gradient estimate $\Delta_T = (\Delta_{1}, \Delta_{2},
\Delta_{3})$ produced by \pomdpg\ must have a relatively small first
component.  Figure~\ref{fig:callgrad} shows a plot of normalized
$\Delta_T$ as a function of $\beta$, for $T=1,000,000$ (sufficiently
large to ensure low variance in $\Delta_T$) and the starting parameter
setting $\theta=(8, 8, 8)$. From the figure, $\Delta_{1}$ starts at
a high value which explains why \conjgrad\ produces ``always accept''
controllers for $\beta=0.0$, and does not become negative until $\beta
\approx 0.93$, a value for which the variance in $\Delta_T$ even for
moderately large  $T$ is relatively high.

\begin{figure}
\begin{center}
\includegraphics[scale=0.47]{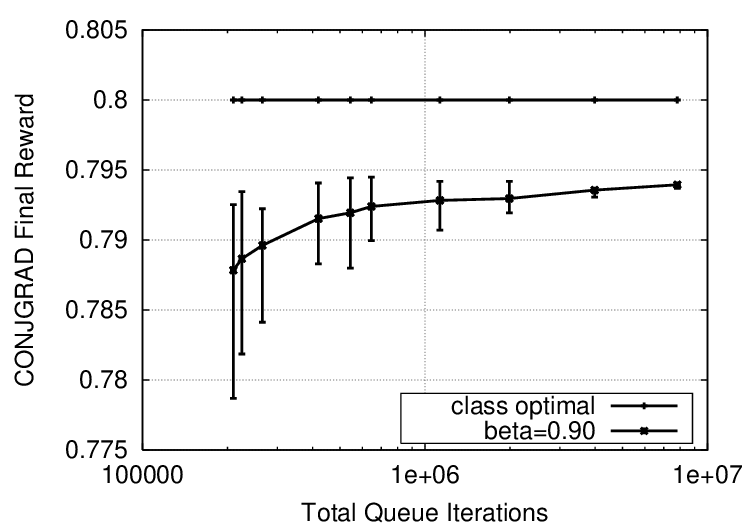}
\includegraphics[scale=0.47]{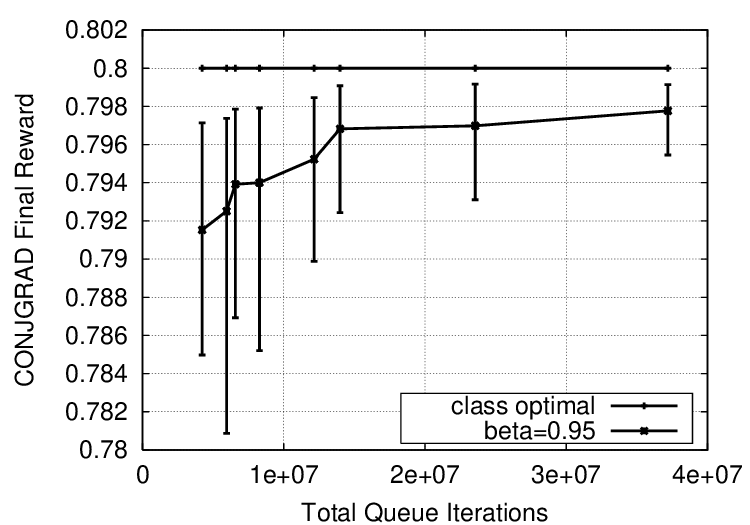}
\caption{Performance of the call admission controller trained by
\conjgrad\, as a function of the total number of iterations of the
queue. The performance was calculated by simulating the controller for
1,000,000 iterations. The graphs are averages from 100 independent runs.
\label{fig:callcg2}}
\end{center}
\end{figure}

A plot of the performance of \conjgrad\ for $\beta=0.9$ and $\beta=0.95$
is shown in Figure \ref{fig:callcg2}. Approximately half of the remaining
2\% in performance can be obtained by setting $\beta=0.9$, while for
$\beta=0.95$ a sufficiently large choice for $T$ gives most of the
remaining performance. For this problem, there is a huge difference
between gaining 98\% of optimal performance, which is achieved for
$\beta=0.0$ and less than 2000 iterations of the queue, and gaining 99\%
of the optimal which requires $\beta=0.9$ and of the order of 500,000
queue iterations. A similar convergence rate and final approximation
error to the latter case were reported for the on-line algorithms by
\citeA[Chapter 7]{marbachthesis98}.

\subsection{Mountainous Puck World}

\begin{figure}
\begin{center}
\includegraphics[scale=0.6]{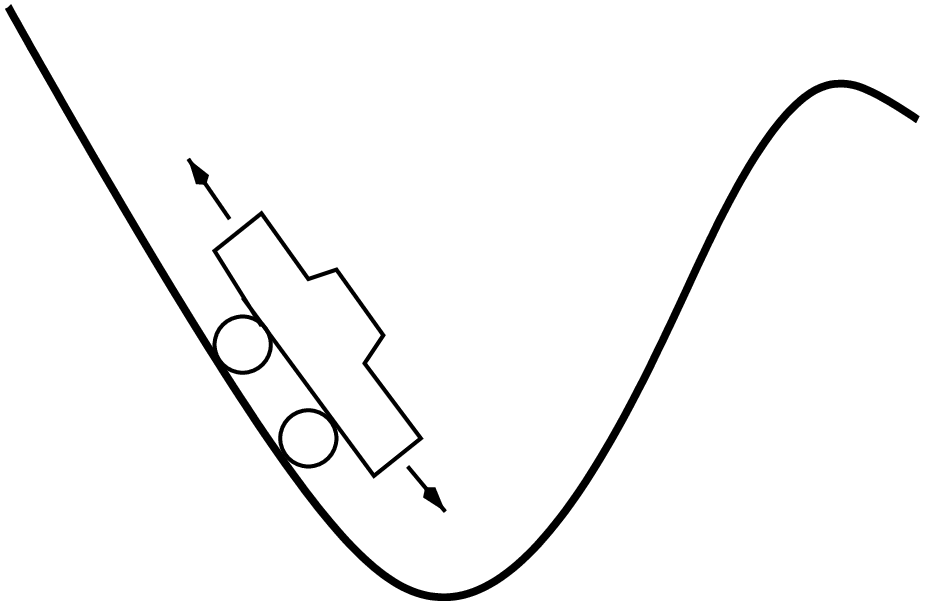}
\caption{The classical ``mountain-car'' task is to apply forward or
reverse thrust to the car to get it over the crest of the hill. The
car starts at the bottom and does not have enough power to drive
directly up the hill.\label{fig:car}}
\end{center}
\end{figure}

The ``mountain-car'' task is a well-studied problem in the
reinforcement learning literature \cite[Example 8.2]{sutton98}. As
shown in Figure~\ref{fig:car}, the task is to drive a car to the
top of a one-dimensional hill. The car is not powerful enough to
accelerate directly up the hill against gravity, so any successful
controller must learn to ``oscillate'' back and forth until it builds
up enough speed to crest the hill. 

\begin{figure}
\begin{center}
\includegraphics[scale=0.55]{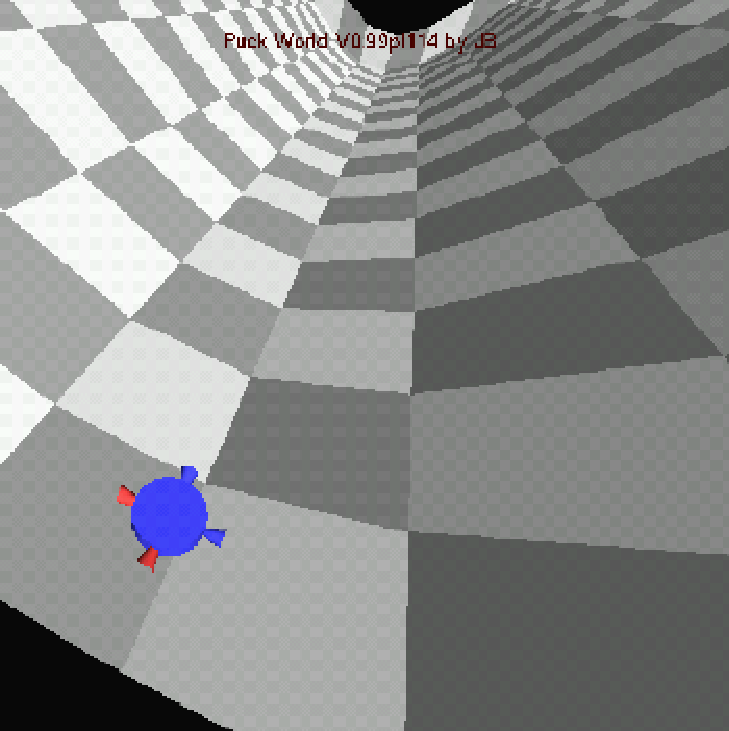}
\caption{In our variant of the mountain-car problem the task is to
navigate a puck out of a valley and onto the northern plateau. The
puck starts at the bottom of the valley and does not have enough power
to drive directly up the hill.\label{fig:mountain}}
\end{center}
\end{figure}

In this section we describe a variant of the mountain car problem
based on the puck-world example of Section~\ref{subsec:puck}. With
reference to Figure~\ref{fig:mountain}, in our problem the task is to
navigate a puck out of a valley and onto a plateau at the northern end
of the valley. As in the mountain-car task, the puck does not have
sufficient power to accelerate directly up the hill, and so has to
learn to oscillate in order to climb out of the valley. Once again we
were able to reliably train near-optimal neural-network controllers
for this problem, using \conjgrad\ and \gsearch, and with \pomdpg\
generating the gradient estimates.

%PB This picture sucks badly. It doesn't add anything but space.

\subsubsection{The World}
The world dimensions, physics, puck dynamics and controls were
identical to the flat puck world described in Section~\ref{subsec:puck},
except that the puck was subject to a constant
gravitational force of $10$ units, the maximum allowed thrust was $3$
units (instead of $5$), and the height of the world varied as follows:
$$
\text{height}(x,y) = 
\begin{cases} 
15 & \text{if $y < 25$ or $y > 75$} \\
7.5  \left[1 - \cos\(\frac{\pi \(\frac{y}{2} - 50\)}{25}\)\right] &
\text{otherwise}. 
\end{cases}
$$
With only $3$ units of thrust, a unit mass puck can not
accelerate directly out of the valley. 

Every 120 (simulated) seconds, the puck was initialized with zero
velocity at the bottom of the valley, with a random $x$ location. 
The puck was given no reward while in the valley or on the
southern plateau, and a reward of $100 - s^2$ while on the northern
plateau, where $s$ was the speed of the puck. We found the speed
penalty helped to improve the rate of convergence
of the neural network controller. 

\subsubsection{The controller} 

After some experimentation we found that a neural-network controller
could be reliably trained to navigate to the northern plateau, or to
stay on the northern plateau once there, but it was difficult to
combine both in the same controller (this is not so surprising since
the two tasks are quite distinct). To overcome this problem, we
trained a ``switched'' neural-network controller: the puck used one
controller when in the valley and on the southern plateau, and then
switched to a second neural-network controller while on the northern
plateau. Both controllers were one-hidden-layer neural-networks with
nine input nodes, five hidden nodes and four output nodes. The nine
inputs were the normalized ($[-1,1]$-valued) $x$, $y$ and $z$ puck
locations, the normalized $x$, $y$ and $z$ locations relative to
center of the northern wall, and the $x$, $y$ and $z$ puck velocities.
The four outputs were used to generate a policy in the same fashion as
the controller of Section~\ref{sec:puckcontrol}. 

An approach requiring less prior knowledge would be to have a third
controller that stochastically selects the base neural network
controller as a function of the puck's location. This ``master''
controller could itself be parameterized and have its parameters
trained along with the base controllers.

\subsubsection{Conjugate gradient ascent}

The switched neural-network controller was trained using the same
scheme discussed in Section~\ref{sec:puckgrad}, except this time the
discount factor $\beta$ was set to $0.98$. 

\begin{figure}
\begin{center}
\includegraphics[scale=0.8]{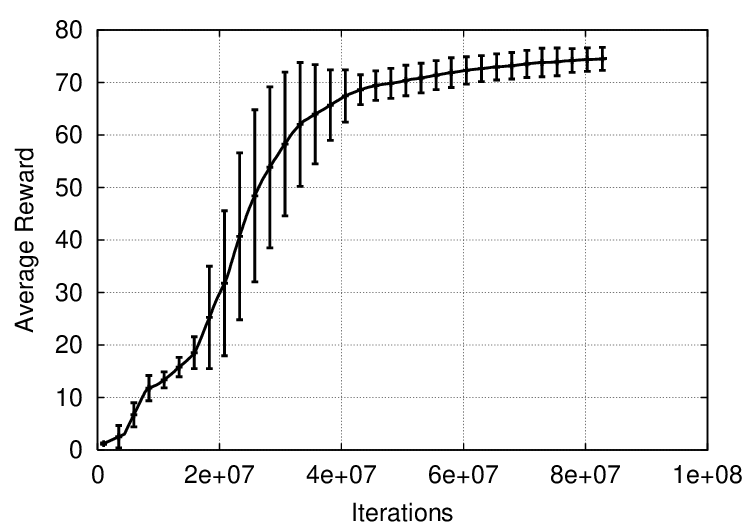}
\caption{Performance of the neural-network puck controller as a
function of the number of iterations of the mountainous puck world, when trained
using \conjgrad. Performance estimates were generated by simulating
for $1,000,000$ iterations. Averaged over 100 independent runs.\label{fig:mountperform}} 
\end{center}
\end{figure}

A plot of the average reward of the neural-network controller is shown
in Figure~\ref{fig:mountperform}, as a function of the number of
iterations of the \pomdp. The graph is an average over 100 independent
runs, with the neural-network controller parameters initialized
randomly in the range $[-0.1,0.1]$ at the start of each run. In this
case no run failed to converge to near-optimal performance.  From the
figure we can see that the puck's performance is nearly optimal
after about 40 million total iterations of the puck world. Although
this figure may seem rather high, to put it in some perspective note that 
a random neural-network controller takes about 10,000 iterations to
reach the northern plateau from a standing start at the base of the
valley. Thus, 40 million iterations is equivalent to only about 4,000 trips 
to the top for a random controller. 

Note that the puck converges to a final average performance around 75, 
which indicates it is spending at least 75\% of its time on the
northern plateau. Observation of the puck's final behaviour shows it
behaves nearly optimally in terms of oscillating back and forth to 
get out of the valley.

\subsection{Choosing $\mathbold{\beta}$ and the Running Time of
$\mathbold{\gpomdp}$}

One aspect of these experiments that required some measure of tuning
is the choice of the $\beta$ parameter and running time $T$ used by
\pomdpg. Although these were selected by trial and error, we have had
some success recently with a scheme for automatically choosing these
parameters as follows. Before any training begins, \gpomdp\ is run for a large number of
iterations whilst simultaneously generating gradient estimates for a
number of different choices of $\beta$. This can be done from a single
simulation simply by maintaining a separate eligibility trace
$z_t$ for each value of $\beta$. Since the bias reduces with increasing
$\beta$, the largest $\beta$ that gives a reasonably low-variance
gradient estimate at the end of the long run is selected as a
``reference'' $\beta$ (the variance is estimated by comparing gradient
estimates at reasonably well-separated intervals towards the end of
the run). Furthermore, since the variance of the gradient estimate
decreases as $\beta$ decreases, all gradient estimates for values of
$\beta$ smaller than the reference $\beta$ will typically have smaller
variance than that of the reference $\beta$. Hence, we can reliably
compare the directions for smaller $\beta$'s with the direction given
by the reference $\beta$, and choose the {\em smallest} $\beta$ whose
corresponding direction is sufficiently close to the reference
$\beta$ direction. We take``sufficiently close'' to mean
within $10^\circ$--$15^\circ$. 

Note that this scheme only works if the original run is sufficiently
long to get a low-variance direction estimate at the right value of
$\beta$. If the right value of $\beta$ is too large then any fixed
bound on the run length can  be made to fail, but this will be a
problem for all algorithms that automatically choose $\beta$. 

Once a suitable $\beta$ has been found, we can go back and find the
point in the original long run where the direction estimate
corresponding to that value of $\beta$ ``settled down'' (again, we
measure the variance of the estimates by sampling at suitably large
intervals, and choose a point where the variance falls below some
chosen value). This time is then used as the running time $T$ for
\gpomdp\ when estimating the gradient direction.  Finally, the running
time used in \gpomdp\ when bracketing the maximum in \gsearch\ can
also be automatically tuned by starting with an initial fixed running
time that is a fraction of $T$, and then continuing until the sign of
the inner product of the estimates produced by \pomdpg\ with the
search direction ``settles down''. With this technique, the sign estimation 
time is usually considerably smaller than the gradient direction estimation
time. 

Another useful heuristic is to re-estimate $\beta$ and \gpomdp's
running time $T$ whenever the parameters $\theta$ change by a large
amount, since a large change in $\theta$ can lead to significant
changes in the mixing time of the \pomdp.

\section{Conclusion}
\label{section:conc}

This paper showed how to use the performance gradient estimates
generated by the \pomdpg\ algorithm \cite{jair_01a} to optimize the
average reward of parameterized \pomdps. We described both a
traditional ``on-line'' stochastic gradient algorithm and an
``off-line'' approach that relied on the use of \gsearch, a robust
line-search algorithm that uses gradient estimates, rather than value
estimates, to bracket the maximum. The off-line approach in particular
was found to perform well on four quite distinct problems: optimizing
a controller for a three-state \mdp, optimizing a neural-network
controller for navigating a puck around a two-dimensional world,
optimizing a controller for a call admission problem, and optimizing a
switched neural-network controller in a variation of the classical
mountain-car task. One reason for the superiority of the off-line
approach is that by searching for a local maximum at each step it
makes much more aggressive use of the gradient information than does
the on-line algorithm. 

For the three-state \mdp\ and the call-admission problems we were able
to provide graphic illustrations of how the bias and variance of the
gradient estimates $\nb\eta$ can be traded against one another by
varying $\beta$ between $0$ (low variance, high bias) and $1$ (high
variance, low bias).

Relatively little tuning was required to generate these results. In
addition, the controllers operated on direct and simple
representations of the state, in contrast to the more complex
representations usually required of value-function based
approaches. 

It is often the case that value-function methods converge much more
rapidly than their policy-gradient counterparts. This is due to the
fact that they enforce constraints on the value-function. With this in
mind an interesting avenue for further research is Actor-Critic
algorithms \cite{BarSutAnd83,baird98,kimura98a,konda99,sutton99} in
which one attempts to combine the fast convergence of value-functions
with the theoretical guarantees of policy-gradient approaches.

Despite the success of the off-line approach in the experiments
described here, the on-line algorithm has advantages in other
settings. In particular, when it is applied to multi-agent
reinforcement learning, both gradient computations and parameter
updates can be performed for distinct agents without any communication
beyond the global distribution of the reward signal. This idea has led
to a parameter optimization procedure for spiking neural networks, and
some successful preliminary results with network routing
\cite{tr_hebb_99,tr_route_01}.

\subsection*{Acknowledgements}
This work was supported by the Australian Research Council, and
benefited from the comments of several anonymous referees. Most of
this research was performed while the first and second authors were
with the Research School of Information Sciences and Engineering,
Australian National University.

\small
\bibliographystyle{theapa}
\bibliography{bib}
\end{document}